\documentclass[pmlr,twocolumn,10pt]{jmlr} 

\usepackage{booktabs}
\usepackage{siunitx}

\usepackage[switch]{lineno}

\usepackage{xspace} 
\usepackage{xcolor}
\newcommand{\method}{\texttt{SHy}\xspace}
\usepackage{enumitem}
\usepackage{threeparttable}

\theorembodyfont{\upshape}
\theoremheaderfont{\scshape}
\theorempostheader{:}
\theoremsep{\newline}

\jmlrvolume{LEAVE UNSET}
\jmlryear{2025}
\jmlrsubmitted{LEAVE UNSET}
\jmlrpublished{LEAVE UNSET}
\jmlrworkshop{Conference on Health, Inference, and Learning (CHIL) 2025} 

\title[SHy]{Self-Explaining Hypergraph Neural Networks for Diagnosis Prediction}

\author{%
\Name{Leisheng Yu} \Email{leisheng.yu@rice.edu}\\
\addr Rice University, United States
\AND
\Name{Yanxiao Cai} \Email{yc139@rice.edu}\\
\addr Rice University, United States
\AND
\Name{Minxing Zhang} \Email{minxing.zhang@duke.edu}\\
\addr Duke University, United States
\AND
\Name{Xia Hu} \Email{xia.hu@rice.edu}\\
\addr Rice University, United States
}

\begin{document}

\maketitle

\begin{abstract}
The burgeoning volume of electronic health records (EHRs) has enabled deep learning models to excel in predictive healthcare. However, for high-stakes applications such as diagnosis prediction, model interpretability remains paramount. Existing deep learning diagnosis prediction models with intrinsic interpretability often assign attention weights to every past diagnosis or hospital visit, providing explanations lacking flexibility and succinctness. In this paper, we introduce \method, a \underline{s}elf-explaining \underline{hy}pergraph neural network model, designed to offer personalized, concise and faithful explanations that allow for interventions from clinical experts. By modeling each patient as a unique hypergraph and employing a message-passing mechanism, \method captures higher-order disease interactions and extracts distinct temporal phenotypes as personalized explanations. It also addresses the incompleteness of the EHR data by accounting for essential false negatives in the original diagnosis record. A qualitative case study and extensive quantitative evaluations on two real-world EHR datasets demonstrate the superior predictive performance and interpretability of \method over existing state-of-the-art models. The code is available at \url{https://github.com/ThunderbornSakana/SHy}.
\end{abstract}

\paragraph*{Data and Code Availability}
This study employs the MIMIC-III \citep{johnson2016mimic} and MIMIC-IV \citep{johnson2023mimic} datasets, both available on the PhysioNet repository \citep{johnson2016physionet}.

\paragraph*{Institutional Review Board (IRB)}
This research does not require IRB approval.

\section{Introduction}
\label{sec:intro}
Electronic health records (EHRs) are large-scale chronologies of patients' hospital visits, encapsulating their longitudinal healthcare experiences \citep{ehr:11}. The surge in EHR data has fostered the development of deep learning models for tasks like diagnosis prediction, mortality prediction, and drug recommendation \citep{ehr:2, other:8, other:7}. Among these, diagnosis prediction based on longitudinal EHRs is of particular significance, as it directly correlates with health risk identification and the quality of personalized healthcare \citep{other:9, van2021clinical}. As shown in \figureref{fig:demo}, longitudinal EHR data comprise sequentially time-stamped hospital visits, each being an unordered set of diagnoses \citep{ehr:8}. The aim of diagnosis prediction is to forecast potential diagnoses for a patient's subsequent visit based on historical records.

\begin{figure}[htbp]
\floatconts
  {fig:demo}
  {\caption{An illustration of diagnosis prediction using the longitudinal EHR of a patient. Diagnoses are denoted by ICD-9 codes.}}
  {\includegraphics[width=\linewidth]{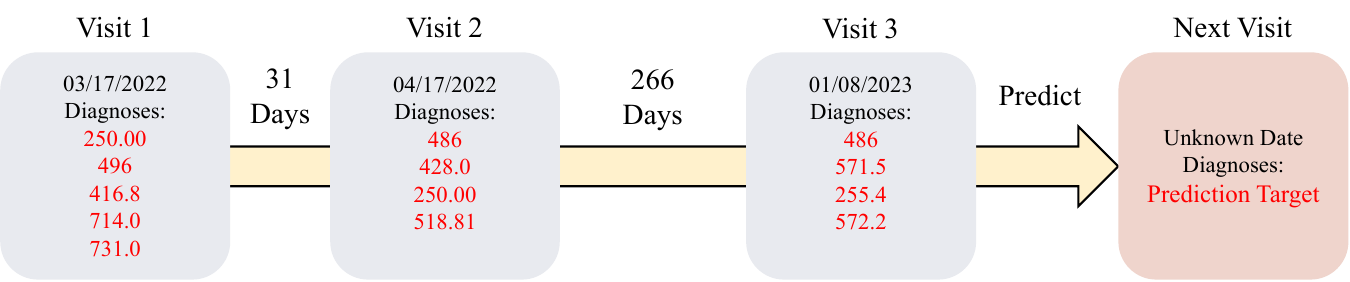}}
\end{figure}

Because the outcomes of diagnosis prediction are communicated to both clinicians and patients, model interpretability is crucial \citep{ehr:23, other:10}. Concerns that post-hoc explanations for black-box deep learning models may not faithfully reflect their actual reasoning processes \citep{ii:2} have spurred research into inherently interpretable models for high-stakes EHR-based applications. These models provide explanations primarily by assigning an attention weight to each historical diagnosis or visit \citep{ehr:12}. This method of explanation has four major limitations:
\begin{itemize}[leftmargin=*]
  \item \textbf{Insufficient personalization:} patients receive a uniform explanation format with importance weights attached to every diagnosis or visit, making it harder for clinicians to discern individual health conditions at first glance compared to customized formats that directly visualize individual comorbidity patterns.
  \item \textbf{Lack of conciseness:} with each diagnosis receiving a weight, size of the explanation equals the input size, failing to filter out redundant information.
  \item \textbf{Not robust against false negative diagnoses:} since patients may not exhibit symptoms for certain chronic conditions, a visit record may be incomplete. Limiting explanations to documented diagnoses can miss false negatives critical for the final prediction.
  \item \textbf{Difficulty in intervention:} models that do not base predictions solely on learned attention weights hinder domain experts from directly influencing the final prediction by editing explanations.
\end{itemize}

To address these challenges, we introduce \method, utilizing temporal phenotypes extracted from the diagnosis history as explanations. Temporal phenotypes, representing observable, informative, and interpretable patterns, can depict the progression of patients' health conditions \citep{ehr:16}. Different from \citet{using:1}, which treats each patient as a temporal graph, \method represents each patient as a hypergraph and extracts sub-hypergraphs as temporal phenotypes, greatly reducing the space complexity. Personalization is enhanced through employing hypergraph neural networks to model higher-order disease interactions within and across visits, leading to tailored disease embeddings that capture individual comorbidity patterns. Moreover, inspired by existing work on hypergraph structure learning \citep{using:3}, \method adds additional diagnosis-visit pairs to the original hypergraph based on embedding similarity, before extracting a collection of phenotypes using the Gumbel-Softmax trick \citep{using:7, using:11}. By enforcing novel regularization, \method produces succinct, non-overlapping, and faithful temporal phenotypes, each reflecting an essential aspect of a patient's evolving health status. Instead of receiving the identical explanation format, different patients now have distinct sets of temporal phenotypes as customized explanations. Since the extracted phenotypes are the basis for the subsequent prediction process, \method is a concept bottleneck model \citep{using:6}, allowing domain experts to intervene in the generated explanations to optimize performance or study model behaviors.

We quantitatively assess both predictive performance and explanation quality, demonstrating that \method generates accurate predictions while providing superior explanations. The main contributions of our work are as follows:
\begin{itemize}[leftmargin=*]
  \item We propose \method, a self-explaining diagnosis prediction model that extracts personalized temporal phenotypes as explanations. This concept bottleneck model, representing patients as hypergraphs, enhances robustness against false negatives and allows domain experts to edit explanations.
  \item \method introduces a novel combination of objectives that ensures the extracted phenotypes are concise, non-overlapping, and faithful.
  \item We perform extensive experiments on two real-world EHR datasets, validating \method's superior predictive performance and interpretability. A case study with clinical experts shows how the explanations generated by \method can be edited.
\end{itemize}

\section{Related Work}

\subsection{Deep Learning on Longitudinal EHRs}
Applying deep learning models to longitudinal EHRs for predictive tasks centers around learning adequate patient representation \citep{ehr:3, baseline:16, tan2024enhancing}. To model temporal disease progression patterns, Doctor AI \citep{baseline:1} leveraged recurrent neural networks (RNNs), StageNet \citep{baseline:4} employed a stage-aware long short-term memory (LSTM) module, and Hi-BEHRT \citep{ehr:18} adopted a hierarchical Transformer effective for long visit histories. To account for irregular time gaps between visits, variants of LSTM and self-attention mechanisms that consider timestamps have been introduced \citep{baseline:9, baseline:11, ehr:20}. To utilize external medical knowledge, methods such as GRAM \citep{baseline:6, baseline:13, baseline:15, ehr:1, baseline:17} infused information from medical ontologies into representation learning via attention mechanisms; PRIME \citep{ehr:21} incorporated rule-based prior medical knowledge through posterior regularization; and SeqCare \citep{ehr:2} employed online medical knowledge graphs with adaptive graph structure learning. To address low-quality data, methods such as GRASP \citep{ehr:22}, which utilized knowledge from similar patients through graph neural networks, and MedSkim \citep{ehr:5}, which filtered out noisy diagnoses using the Gumbel-Softmax trick, were developed. Existing approaches tried to enhance model interpretability by giving weights to every past diagnosis or visit: methods such as RETAIN achieved this with attention mechanisms \citep{baseline:2, baseline:3, baseline:5, baseline:12, ehr:13, ehr:19, baseline:14, ehr:7}; AdaCare \citep{baseline:8} employed a scale-adaptive feature recalibration module. To mitigate data insufficiency in certain scenarios, \citet{ehr:6} pre-trained BERT on a large EHR corpus and fine-tuned it on smaller datasets.

\subsection{Intrinsically Interpretable Models}
Intrinsic interpretability, which means that predictive models provide their own explanations, is favored over post-hoc interpretability that necessitates a separate model for explaining a black-box model, because the explanations provided by intrinsically interpretable models are exploited in the decision-making process and faithful to model predictions \citep{ii:1, ii:2}. SENN \citep{using:4} is a class of self-explaining neural networks whose interpretability is enforced via regularization. A self-explaining deep learning model proposed by \citet{using:5} utilizes an autoencoder and a prototype classifier network to provide case-based rationales. The attention mechanism has been employed to achieve intrinsic interpretability by using attention weights as explanations \citep{ii:6}, but its reliability is arguable \citep{ii:7, ii:10, ii:11}. SITE \citep{ii:4} emphasizes robust interpretations equivariant to geometric transformations. ProtoVAE \citep{ii:5} leverages a variational autoencoder to learn class-specific prototypes. The Bort optimizer \citep{ii:9} enhances model explainability by imposing boundedness and orthogonality constraints on model parameters. Intrinsic interpretability has been studied in healthcare \citep{ehr:17}. However, unlike \method, existing self-explaining deep learning models for EHR-based diagnosis prediction fail to provide temporal explanations that reflect the distinct comorbidities of individual patients.

The related work on deep learning on hypergraphs is discussed in \appendixref{apd:first}.

\section{Methodology}
\label{sec:method}
An overview of the proposed model, \method, is illustrated in \figureref{fig:overview}. Before delving into the details of the method, it is essential to formulate the problem of diagnosis prediction using longitudinal EHR data.

\begin{figure}[ht]
\floatconts
  {fig:overview}
  {\caption{Overview of \method.}}
  {\includegraphics[width=\linewidth]{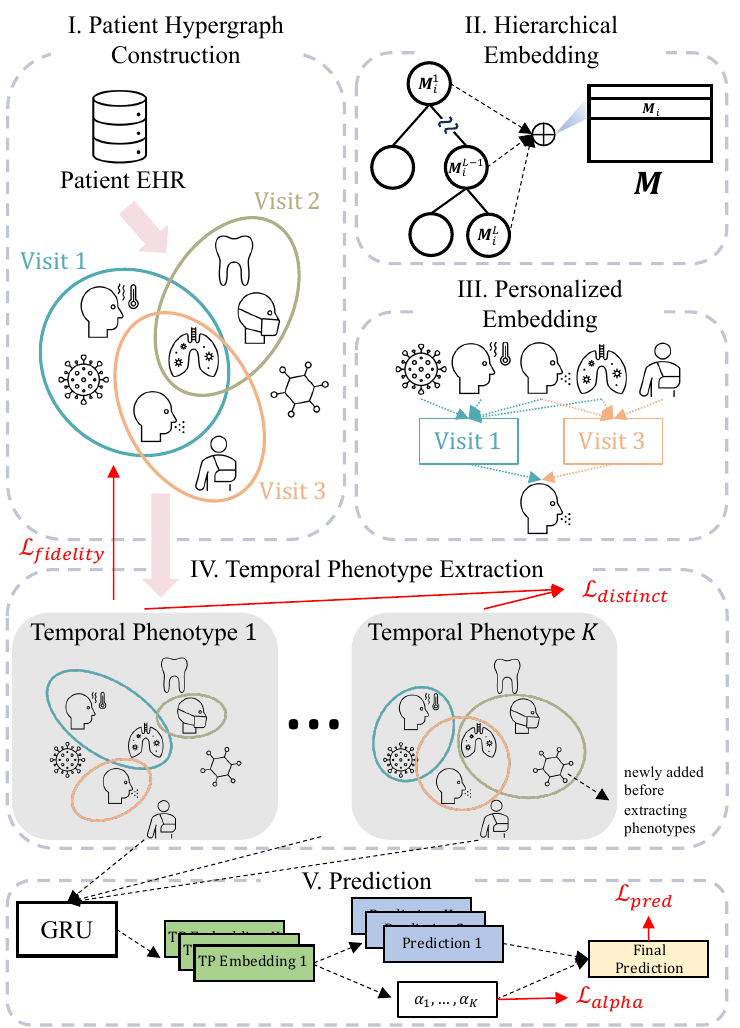}}
\end{figure}

\subsection{Problem Formulation}
An EHR dataset contains various diseases, each assigned a diagnosis code according to the International Classification of Diseases, Ninth Revision (ICD-9). ICD-9 codes have a hierarchical structure. For example, both \textit{acute respiratory failure} (ICD-9 code $518.81$), and \textit{chronic respiratory failure} ($518.83$) are children of \textit{other diseases of lung} ($518.8$). The set of all unique diagnosis codes within an EHR dataset is denoted as $c_{1}, c_{2}, ..., c_{|\set{C}|} \in \set{C}$, and $|\set{C}|$ indicates the size of this set. Assuming that we have $N$ patients in a longitudinal EHR dataset, we can represent the $n$-th patient as a sequence, $[(\vec{e_{1}^{n}}, t_{1}^{n}), (\vec{e_{2}^{n}}, t_{2}^{n}), ..., (\vec{e_{T^{(n)}}^{n}}, t_{T^{(n)}}^{n})]$, where $T^{(n)}$ is the number of past visits, $\vec{e_{j}^{n}}$ denotes the diagnosis codes recorded in the $j$-th visit, and $t_{j}^{n}$ signifies the timestamp. $\vec{e_{j}^{n}} \in \{0, 1\}^{|\mathcal{C}|}$ is a multi-hot vector, where the $i$-th entry equals $1$ if visit $j$ includes $c_{i}$. The task of diagnosis prediction is to predict the diagnoses in the next visit, $\vec{e_{T^{(n)} + 1}^{n}}$, based on past visits. To simplify notation, the superscript $n$ is dropped in later sections, where the context allows.

\subsection{Personalized Disease Representation Learning}
Before detailing how \method learns personalized diagnosis code embeddings to reflect different comorbidities, we outline how it models hierarchical disease relationships, a common practice in related work \citep{ehr:1, ehr:4, baseline:13, baseline:15}.

\subsubsection{Learning Hierarchical Disease Embeddings}
ICD-9 organizes diseases into a tree structure with each child node having a single parent. This medical ontology can be used to enhance the representation learning process for diagnosis codes, defining relative disease distances. Thus, previous work has explored effective modeling of this hierarchical relationship, a type of external medical knowledge \citep{baseline:6, baseline:14, baseline:17}. As the primary focus of our work is not developing new strategies to utilize ICD-9, we adopt the hierarchical embedding module from CGL \citep{baseline:7}. Specifically, the ICD-9 tree comprises $L$ levels (we set $L=4$ following CGL), and most diagnosis codes in the EHR dataset are leaf nodes. We accommodate non-leaf nodes in the dataset by creating virtual child nodes, and padding them to level $L$. Each node at every level gets an initialized embedding, resulting in $L$ embedding matrices, $\{\mathbf{M}^{1}, \mathbf{M}^{2}, ..., \mathbf{M}^{L}\}$. Here, $\mathbf{M}^{l} \in \mathbb{R}^{n_l \times d_{c}}$ is the embedding matrix for the $l$-th level, with $n_l$ and $d_{c}$ denoting the number of nodes at the level $l$ and the embedding dimension, respectively. Then, for example, the ontology-aware embedding, $\mathbf{M}_{i} \in \mathbb{R}^{Ld_{c}}$, for $c_{i}$ is obtained via concatenating its embeddings with those of its ancestors:
\begin{equation} \label{eq:1}
\mathbf{M}_{i} = \mathbf{M}_{i}^{1}\mathbin\Vert\mathbf{M}_{i}^{2}\mathbin\Vert...\mathbin\Vert\mathbf{M}_{i}^{L}.
\end{equation}
Simply put, after this stage, an embedding table, $\mathbf{M} \in \mathbb{R}^{|\set{C}| \times Ld_{c}}$, is initialized as learnable parameters for all diagnosis codes in the EHR dataset.

\subsubsection{Constructing Patient Hypergraphs}
To model individual comorbidities for personalized patient representation, we should represent each patient as an independent entity and effectively capture disease interactions. One intuitive method is to represent patients as ordinary graphs where diseases form nodes and co-occurrences create edges. However, this strategy requires the construction of an adjacency matrix of size $|\set{C}| \times |\set{C}|$ for every patient. Since an EHR dataset typically contains thousands of unique diagnosis codes and tens of thousands of patients, this leads to prohibitively high space complexity. Moreover, an ordinary graph only models pairwise interactions, but the co-occurrence of multiple diseases in a single visit suggests non-pairwise relationships. Hypergraphs, as an alternative to ordinary graphs, can address these issues. They can represent higher-order relations because hyperedges in hypergraphs can connect any number of nodes. Therefore, we model each patient as a hypergraph, with diseases as nodes and hospital visits as hyperedges. In this way, higher-order interactions among diseases can be captured. Each patient hypergraph, $\mathcal{G} = (\set{C}, \set{E})$, can be represented as an incidence matrix $\mathbf{P} \in \{0, 1\}^{|\set{C}| \times T}$, where $\mathbf{P}_{ij} = 1$ if the $j$-th visit contains $c_{i}$, and each hyperedge $\vec{e_{j}} \in \set{E}$ is a subset of $\set{C}$. Since the average number of hospital visits of patients in typical EHRs is below $10$, modeling patients as hypergraphs greatly reduces space complexity.

\subsubsection{Modeling Individual Comorbidities}
To model distinctive comorbidities and thus learn personalized diagnosis code embeddings, \method conducts message passing on the constructed patient hypergraphs. With numerous hypergraph neural network architectures in the existing literature, we experimented with several state-of-the-art models, the results of which are discussed in \appendixref{apd:second}. We empirically selected UniGIN \citep{using:8} as \method's message passing mechanism. UniGIN generalizes Graph Isomorphism Networks (GIN) \citep{other:5} to hypergraphs by formulating it as a two-stage aggregation process. Specifically, \method first obtains visit embeddings by aggregating the embeddings of diagnosis codes within the respective visit:
\begin{equation} \label{eq:2}
\mathbf{V}^{(z)}_{j} = \frac{1}{|\vec{e_{j}}|}\sum_{i \in \vec{e_{j}}} \mathbf{M}^{(z)}_{i},
\end{equation}
where $\mathbf{V}^{(z)}_{j}$, $|\vec{e_{j}}|$, and $z$ denote the embedding of the $j$-th visit, number of diseases within the $j$-th visit, and the index of UniGIN layers, respectively. We set $\mathbf{M}^{(0)}_{i} = \mathbf{M}_{i}$. Next, \method updates diagnosis embeddings by aggregating the embeddings of visits containing the corresponding diagnosis:
\begin{equation} \label{eq:3}
\mathbf{M}^{(z+1)}_{i} = \sigma(\mathbf{W}_{\textrm{UniGIN}}^{(z)}((1 + \varepsilon)\mathbf{M}^{(z)}_{i} + \sum_{j \in \set{E}_{i}} \mathbf{V}^{(z)}_{j})),
\end{equation}
where $\sigma$ is Leaky ReLU, $\mathbf{W}_{\textrm{UniGIN}}^{(z)} \in \mathbb{R}^{d^{(z+1)}_{c} \times d^{(z)}_{c}}$ denotes learnable weights of the $z$-th UniGIN layer, $\varepsilon$ is a learnable parameter, and $\set{E}_{i}$ represents the set of indices of visits including $c_{i}$. By stacking $Z$ layers of message passing mechanisms on individual patient hypergraphs, \method adeptly models complex interactions among diseases within the $Z$-hop neighborhood. Thus, disease embeddings for a specific patient are strongly influenced by the combinations of diseases that this patient had within and across visits.

In summary, after this stage, an updated personalized embedding table of diagnosis codes, $\mathbf{M}^{(Z)} \in \mathbb{R}^{|\set{C}| \times d^{(Z)}_{c}}$, is obtained for each patient. The superscript $(Z)$ will be omitted in subsequent sections.

\subsection{Temporal Phenotype Extraction \& Modeling}
Since the diagnosis history of a patient possibly suffers from incompleteness, \method introduces false negative disease-visit pairs into the constructed patient hypergraph prior to phenotype extraction. These additions are based on the multi-head weighted cosine similarity between nodes and hyperedges:
\begin{equation} \label{eq:4}
\mathbf{S}_{ij} = \frac{1}{n_{s}}\sum_{k=1}^{n_{s}} \frac{(\mathbf{\Phi}_{k} \odot \mathbf{M}_{i}) \cdot (\mathbf{\Phi}_{k} \odot \mathbf{V}_{j})}{\|\mathbf{\Phi}_{k} \odot \mathbf{M}_{i}\|\|\mathbf{\Phi}_{k} \odot \mathbf{V}_{j}\|},
\end{equation}
where $\mathbf{S}_{ij}$ denotes the similarity score between the embeddings of $c_{i}$ and the $j$-th visit, $\mathbf{\Phi} \in \mathbb{R}^{n_{s} \times d^{(Z)}_{c}}$ represents a stack of $n_{s}$ independent learnable weight vectors, and $\mathbf{V}_{j}$ is obtained through \equationref{eq:2}. To avoid adding redundant connections, \method enforces $\mathbf{S}_{ij} = 0$ if $c_{i}$ was originally included in the $j$-th visit. Next, \method derives a supplementary hypergraph, $\Delta\mathbf{P} \in \{0, 1\}^{|\set{C}| \times T}$, based on the similarity scores:
\begin{equation} \label{eq:5}
\Delta\mathbf{P}_{ij} =
    \begin{cases}
      1, &\mathbf{S}_{ij} \in \texttt{topk}(\mathbf{S}, p|\sum_{j=1}^{T} \vec{e_{j}}|)\\
      0, &\mathbf{S}_{ij} \notin \texttt{topk}(\mathbf{S}, p|\sum_{j=1}^{T} \vec{e_{j}}|),
    \end{cases} 
\end{equation}
where $p$ is the ratio of the number of connections in the supplementary hypergraph to those in the initial patient hypergraph. Then, an updated patient hypergraph, $\tilde{\mathbf{P}} = \mathbf{P} + \Delta\mathbf{P}$, is produced. \method essentially augments the original patient records with additional disease-visit pairs possessing the highest similarity scores.

\method extracts $K$ temporal phenotypes, each being a unique subgraph of the updated patient hypergraph. Specifically, to extract Phenotype $k$, \method learns a binary matrix denoted by $\mathbf{\Gamma}^{k} \in \{0, 1\}^{|\set{C}| \times T}$. Each entry, $\mathbf{\Gamma}^{k}_{ij}$, serves as a masking factor for $\tilde{\mathbf{P}}_{ij}$ and is a random variable following a Bernoulli distribution parameterized by a probability weight, $\mathbf{O}^{k}_{ij}$. This weight is derived from the embeddings of the corresponding disease and visit:
\begin{equation} \label{eq:6}
\mathbf{O}^{k}_{ij} = \textrm{MLP}([\mathbf{M}_{i}\mathbin\Vert\mathbf{V}_{j}]).
\end{equation}
To allow backpropagation while producing the discrete binary matrix $\mathbf{\Gamma}^{k}$, \method employs the Gumbel-Softmax trick:
\begin{equation} \label{eq:7}
\mathbf{\Gamma}^{k}_{ij} = \sigma(\frac{\log(\frac{\mathbf{O}^{k}_{ij}}{1 - \mathbf{O}^{k}_{ij}}) + (\delta^{0} - \delta^{1})}{\tau}),
\end{equation}
where $\delta^{0}, \delta^{1} \sim \textrm{Gumbel}(0,1)$, $\sigma$ is the sigmoid function, and $\tau$ is the temperature. With the generated binary matrix, \method extracts Phenotype $k$ through
\begin{equation} \label{eq:8}
\mathbf{\Psi}^{k} = \tilde{\mathbf{P}} \odot \mathbf{\Gamma}^{k},
\end{equation}
where $\mathbf{\Psi}^{k} \in \{0, 1\}^{|\set{C}| \times T}$ is the incidence matrix of Phenotype $k$. The process, from \equationref{eq:6} to \equationref{eq:8}, constitutes a temporal phenotype extractor. \method utilizes $K$ independent temporal phenotype extractors, operating concurrently to generate $\{\mathbf{\Psi}^{1}, \mathbf{\Psi}^{2}, ..., \mathbf{\Psi}^{K}\}$.

\subsection{Prediction and Objectives}
To predict subsequent diagnoses, \method embeds $K$ temporal phenotypes using a Gated Recurrent Unit (GRU) and location-based attention mechanism:
\begin{align}
\mathbf{V}^{k} &= {\mathbf{\Psi}^{k}}^{\!\!\top}\mathbf{M} \label{eq1}\\ 
\mathbf{H}^{k}_{1}, \mathbf{H}^{k}_{2}, ..., \mathbf{H}^{k}_{T} &= \textrm{GRU}(\mathbf{V}^{k}_{1}, \mathbf{V}^{k}_{2}, ..., \mathbf{V}^{k}_{T}) \label{eq2}\\
\boldsymbol{\alpha}^{k} &= \textrm{Softmax}(\textrm{MLP}(\mathbf{H}^{k})) \label{eq3}\\
\mathbf{U}_{k} &= \boldsymbol{\alpha}^{k}\mathbf{H}^{k}\label{eq4},
\end{align}
where $\mathbf{V}^{k} \in \mathbb{R}^{T \times d^{(Z)}_{c}}$ is a stack of embeddings of visits in Phenotype $k$, $\mathbf{H}^{k}_{t} \in \mathbb{R}^{d_{hid}}$ denotes the hidden state corresponding to the $t$-th visit, $\boldsymbol{\alpha}^{k} \in \mathbb{R}^{T}$ represents weights of hidden states, and $\mathbf{U}_{k} \in \mathbb{R}^{d_{hid}}$ is the embedding of Phenotype $k$. Next, \method derives importance weights, $\boldsymbol{\alpha} \in \mathbb{R}^{K}$, for all phenotypes with a self-attention mechanism:
\begin{align}
head_{i} &= \textrm{Attention}(\mathbf{U}\mathbf{W}^{Q}_{i}, \mathbf{U}\mathbf{W}^{K}_{i}, \mathbf{U}\mathbf{W}^{V}_{i}) \label{eq5}\\ 
\boldsymbol{\alpha} &= \textrm{Softmax}([head_{1} \mathbin\Vert head_{2} \mathbin\Vert ... \mathbin\Vert head_{n_{h}}]\mathbf{w}^{O})\label{eq6}.
\end{align}
Here, $\mathbf{W}^{Q}_{i} \in \mathbb{R}^{d_{hid} \times d_{Q}}$, $\mathbf{W}^{K}_{i} \in \mathbb{R}^{d_{hid} \times d_{Q}}$, $\mathbf{W}^{V}_{i} \in \mathbb{R}^{d_{hid} \times d_{V}}$, and $\mathbf{w}^{O} \in \mathbb{R}^{d_{V}}$ are learnable parameters, and $\textrm{Attention}(\mathbf{Q}, \mathbf{K}, \mathbf{V}) = \textrm{Softmax}(\frac{\mathbf{Q}\mathbf{K}^{\!\!\top}}{\sqrt{d_{K}}})\mathbf{V}$, where $d_{K}$ denotes the dimensionality of $\mathbf{K}$. The final prediction, $\vec{\hat{y}} \in \mathbb{R}^{|\set{C}|}$, is a weighted sum of predictions from $K$ phenotypes:
\begin{equation} \label{eq:9}
\vec{\hat{y}} = \boldsymbol{\alpha}(\textrm{Softmax}(\mathbf{U}\mathbf{W} + \vec{b})),
\end{equation}
where $\mathbf{W} \in \mathbb{R}^{d_{hid} \times |\set{C}|}$ and $\mathbf{b} \in \mathbb{R}^{|\set{C}|}$ denote learnable parameters. The prediction loss for all patients is calculated using cross-entropy:
\[\mathcal{L}_{pred} = -\frac{1}{N}\sum_{n=1}^{N} (\vec{y_{n}^{\!\!\top}}\log(\vec{\hat{y}_{n}}) + (1-\vec{y_{n}})^{\!\!\top}\log(1-\vec{\hat{y}_{n}})),\]
where $\vec{y_{n}} = \vec{e_{T^{(n)} + 1}^{n}}$. The $K$ temporal phenotypes, along with their respective importance weights, act as model explanations. To ensure that the extracted phenotypes faithfully preserve relevant input information, \method employs a GRU decoder, the details of which are introduced in \appendixref{apd:third}, to reconstruct the original patient hypergraph—a sequence of multi-hot vectors—from the concatenated phenotype embeddings. The fidelity of explanations is enhanced by penalizing the reconstruction error:
\begin{multline*}
\mathcal{L}_{fidelity} = -\frac{1}{N}\sum_{n=1}^{N} \frac{1}{T^{(n)}|\set{C}|} \sum_{j=1}^{T^{(n)}} \sum_{i=1}^{|\set{C}|} (\mathbf{P}^{n}_{ij} \cdot \log\hat{\mathbf{P}}^{n}_{ij} \\ + (1 - \mathbf{P}^{n}_{ij}) \cdot \log(1-\hat{\mathbf{P}}^{n}_{ij})),
\end{multline*}
where $\hat{\mathbf{P}}^{n}$ denotes the reconstructed hypergraph for the $n$-th patient. Moreover, the $K$ temporal phenotypes need to be non-overlapping in order to be meaningful, i.e., different phenotypes include different diagnoses for a given visit. \method achieves distinctiveness by penalizing common diagnoses across $K$ binary masks for one visit:
\begin{align*}
\mathbf{B}_{j} &= [\mathbf{\Gamma}_{\cdot j}^{1}, \mathbf{\Gamma}_{\cdot j}^{2}, ..., \mathbf{\Gamma}_{\cdot j}^{K}] \in \mathbb{R}^{|\set{C}| \times K} \\
\mathcal{L}_{distinct} &= \frac{1}{N}\sum_{n=1}^{N} \frac{1}{T^{(n)}}\sum_{j=1}^{T^{(n)}} \|\mathbf{I}_{T^{(n)}} - {\mathbf{B}^{n}_{j}}^{\!\!\top}\mathbf{B}^{n}_{j}\|_{2},
\end{align*}
where $\mathbf{I}_{T^{(n)}}$ denotes an identity matrix, $\mathbf{\Gamma}_{\cdot j}^{k}$ represents the $j$-th column of the binary mask $\mathbf{\Gamma}^{k}$, and $\mathbf{B}_{j}$ specifies the diagnoses each masking matrix retains for the $j$-th visit. The aim is to encourage columns of $\mathbf{B}_{j}$ to form an orthonormal set. Since $\mathbf{B}_{j}$ is a binary matrix, columns being orthonormal vectors implies that there is no overlap across the $K$ phenotypes for the $j$-th visit and each phenotype only includes one diagnosis for this visit. Thus, $\mathcal{L}_{distinct}$ also pushes the phenotypes to be concise. Lastly, \method enforces regularization to prevent the distribution of the $K$ attention weights from being uniform or too extreme, ensuring clear relative importance among phenotypes and no extracted phenotype receiving $0$ as the weight:
\[\mathcal{L}_{alpha} = -\frac{1}{N}\sum_{n=1}^{N} (\sqrt{\frac{\sum_{k} (\boldsymbol{\alpha}^{n}_{k} - \bar{\boldsymbol{\alpha}^{n}})^{2}}{K}} - \|\boldsymbol{\alpha}^{n}\|_{2}),\]
where $\boldsymbol{\alpha}^{n}$ denotes attention weights for the phenotypes of the $n$-th patient. The final loss function is as follows:
\begin{equation} \label{eq:10}
\mathcal{L} = \mathcal{L}_{pred} + \epsilon \cdot \mathcal{L}_{fidelity} + \eta \cdot \mathcal{L}_{distinct} + \omega \cdot \mathcal{L}_{alpha},
\end{equation}
where $\epsilon$, $\eta$, and $\omega$ are hyperparameters. This weighted combination enables \method to provide distinct and concise temporal phenotypes, with each elucidating a unique aspect of the final prediction faithfully. A salient feature of \method is that it is a concept bottleneck model, meaning that it derives a set of phenotypes and utilizes them to predict the target. This attribute allows clinical practitioners to edit the extracted temporal phenotypes, with changes subsequently propagated to the final prediction via \equationref{eq1} through \equationref{eq:9}.

\section{Experiments}
\label{sec:experiments}
In this section, we evaluate \method against state-of-the-art diagnosis prediction models in terms of predictive performance and interpretability, and assess the effectiveness of its novel components. We further demonstrate \method's robustness against false negative diagnoses and its explanation capabilities, highlighting how domain experts can seamlessly intervene in the prediction process.

\subsection{Data Description \& Experimental Setup}
We utilize the MIMIC-III and MIMIC-IV datasets for our experiments, both of which are de-identified and publicly-available collections of electronic health records associated with patients admitted to the Beth Israel Deaconess Medical Center \citep{other:3, johnson2016mimic}. As observed in \tableref{tab:dataset_info}, MIMIC-IV notably possesses a greater number of diagnosis codes and patients.

Our evaluation metrics for diagnosis prediction are Recall@$k$ and nDCG@$k$. We employ three metrics for a quantitative evaluation of explanation quality. Faithfulness measures the Pearson correlation between importance weights and changes in prediction upon removal of the corresponding explanation units from the input. Ranging from $-1$ to $1$, a high value indicates that the provided explanations can faithfully reflect the model's reasoning process \citep{using:4}. Complexity counts the number of diagnoses in the given explanation: a smaller number implies greater conciseness. Distinctness, exclusive to evaluating \method's explanations, gauges the overlap among temporal phenotypes:
\[\textrm{Distinctness} = \frac{1}{N}\sum_{n=1}^{N} \frac{\|\mathbf{\Psi}^{n,1} + \mathbf{\Psi}^{n,2} + ... + \mathbf{\Psi}^{n,K}\|_{0}}{\|\mathbf{\Psi}^{n,1} + \mathbf{\Psi}^{n,2} + ... + \mathbf{\Psi}^{n,K}\|},\]
where $\mathbf{\Psi}^{n,k}$ denotes the $k$-th phenotype for the $n$-th patient. A higher Distinctness value implies that \method produces more varied phenotypes for an individual patient. Details on baselines and hyperparameters are introduced in \appendixref{apd:fourth}.

\subsection{Predictive Performance Comparison}
For both MIMIC-III and MIMIC-IV, where the average number of diagnoses per visit ranges between $10$ and $20$, we set $k=\{10, 20\}$ for Recall@$k$ and nDCG@$k$. \tableref{tab:predictive_expt} compares the performance of \method with baseline models for diagnosis prediction, along with the count of learnable parameters for each model. Overall, \method delivers strong predictive performance, surpassing most baselines across all metrics on both datasets. In particular, \method outperforms ConCare and Doctor AI, two black-box models, by up to $37.44\%$ in Recall@$10$ (MIMIC-III) and $64.43\%$ in nDCG@$20$ (MIMIC-IV), respectively. In contrast, \method's edge over T-LSTM isn't as pronounced, especially on MIMIC-IV, which can be attributed to T-LSTM's much larger number of parameters. The large model size of T-LSTM can bring high expressiveness given ample training data. When juxtaposed with interpretable models, \method's superiority is even more pronounced: outperforming RETAIN, Dipole, and Timeline by up to $70.53\%$ in Recall@$20$ on MIMIC-IV, $66.84\%$ in nDCG@$10$ on MIMIC-IV, and $28.50\%$ in nDCG@$10$ on MIMIC-III, respectively. It is intriguing to note that \method's margin over AdaCare is much narrower ($4.39\%$ in Recall@$20$ on MIMIC-III), with the difference reducing further on MIMIC-IV. This is potentially due to AdaCare's larger size.

Although \method adopts a hierarchical disease embedding approach from CGL, it outclasses CGL by $5.86\%$ in Recall@$10$ on MIMIC-IV. This can be ascribed to CGL's reliance on a disease co-occurrence graph, which lacks the capability to capture higher-order disease interactions in the manner hypergraphs can. However, modeling patients as hypergraphs is insufficient. The essence lies in performing message passing to learn personalized disease embeddings that reflect individual comorbidities. This claim can be supported by CGL's superior performance over \method w/o. MP across all metrics.

The other interesting observation is that on MIMIC-III, \method outperforms \method($K=1$) by up to $4.61\%$ in Recall@$20$, but on MIMIC-IV, \method($K=1$) surpasses the standard \method in certain metrics. To unravel this discrepancy, we note that \method($K=1$) can only extract one temporal phenotype, rendering the objectives $\mathcal{L}_{distinct}$ and $\mathcal{L}_{alpha}$ inapplicable. Therefore, \method($K=1$) is trained with only two objectives, $\mathcal{L}_{pred}$ and $\mathcal{L}_{fidelity}$. Similarly, \method outperforms \method  w/o. $\mathcal{L}_{distinct, alpha}$ on MIMIC-III across all metrics, but is surpassed by \method  w/o. $\mathcal{L}_{distinct, alpha}$ on MIMIC-IV. These findings suggest that the regularization we enforce for enhanced interpretability can be particularly beneficial with smaller training datasets, helping the model achieve better performance by mitigating the risk of overfitting. In contrast, with voluminous training data, these added objectives can actually act as a constraint, limiting the predictive power of the model. Supporting this observation, \method(only $\mathcal{L}_{pred}$) consistently, albeit marginally, outshines the standard \method on MIMIC-IV, demonstrating the inherent trade-off between model interpretability and prediction accuracy, while on MIMIC-III their performances are largely indistinguishable.

\begin{table*}[ht]
\floatconts
  {tab:predictive_expt}
  {\caption{Results of Predictive Performance Evaluation across Two EHR Datasets}}
  {
  \begin{threeparttable}
\resizebox{\textwidth}{!}{
    \begin{tabular}{@{}c|cccccccccc@{}}
      \toprule
      Model &
      Recall@$10$ &
      Recall@$20$ &
      nDCG@$10$ &
      nDCG@$20$ &
      \# Params &
      Recall@$10$ &
      \multicolumn{1}{l}{Recall@$20$} &
      \multicolumn{1}{l}{nDCG@$10$} &
      \multicolumn{1}{l}{nDCG@$20$} &
      \multicolumn{1}{l}{\# Params} \\ \midrule
               & \multicolumn{5}{c}{MIMIC-III}                                          & \multicolumn{5}{c}{MIMIC-IV}               \\ \midrule
      Doctor AI  & $0.2154$ & $0.3021$ & $0.3353$ & $0.3282$              & $2.90$M  & $0.1811$     &  $0.2490$    &  $0.2591$      &   $0.2578$     & $4.71$M  \\
      RETAIN     & $0.2024$ & $0.2958$ & $0.3136$ & $0.3167$ & $1.47$M  & $0.1713$ & $0.2504$ & $0.2529$ & $0.2605$ & $2.37$M  \\
      Dipole     & $0.2022$ & $0.2965$ & $0.3130$ & $0.3170$              & $2.80$M  &  $0.1721$      &  $0.2514$      &   $0.2539$     &   $0.2615$     & $4.16$M  \\
      T-LSTM     & $0.2758$ & $0.3735$ & $0.4132$ & $0.4055$              & $30.38$M & $0.3328$     &   $0.4268$     &   $0.4208$     &   $0.4232$     & $48.49$M \\
      GRAM       & $0.2219$       & $0.3308$       & $0.3489$       & $0.3551$       & $1.62$M  &   $0.1796$     &  $0.2515$      &  $0.2542$      &   $0.2611$     & $2.08$M  \\
      CGL        & $0.2698$       & $0.3679$       & $0.4125$       & $0.4022$              & $1.52$M  &  $0.3159$      &   $0.4198$     &    $0.4100$    &   $0.4180$     & $2.90$M  \\
      Timeline   & $0.2153$       & $0.3090$       & $0.3218$       & $0.3239$       & $1.20$M       & $0.2879$       &  $0.3898$    & $0.3651$       &  $0.3748$      &  $1.84$M      \\
      AdaCare    & $0.2662$ & $0.3670$ & $0.3970$ & $0.3935$ & $16.43$M & $0.3323$ & $0.4259$ & $0.4197$ & $0.4221$ & $42.98$M \\
      StageNet   & $0.2174$ & $0.3113$ & $0.3245$ & $0.3270$              & $4.82$M  &  $0.3013$      &    $0.4003$    &  $0.3760$      &   $0.3850$     & $6.63$M  \\
      ConCare    & $0.2019$ & $0.2937$ & $0.3066$ & $0.3103$             & $2.63$M  &  $0.2715$      &   $0.3599$     &   $0.3372$     &   $0.3456$     & $3.19$M  \\
      Chet       & $0.2155$       & $0.3102$       & $0.3259$       & $0.3296$       & $2.12$M  &  $0.1814$      &  $0.2566$      &   $0.2609$     &   $0.2671$     & $3.48$M  \\
      SETOR      & $0.2266$       & $0.3317$       & $0.3503$       & $0.3568$       & $10.02$M &   $0.1810$     &   $0.2532$    &  $0.2556$     &   $0.2648$     & $12.78$M \\
      MIPO       & $0.2251$       & $0.3321$       & $0.3499$       & $0.3589$       & $12.13$M       &  $0.1819$      &  $0.2545$      &   $0.2555$     &  $0.2651$      &  $14.92$M      \\ \midrule
      \method($K=1$)  & $0.2660$   &  $0.3662$  &  $0.4006$  &  $0.3965$  & $2.00$M  & $0.3340$    & $0.4272$       &   $0.4238$     &   \underline{$0.4257$}     & $2.89$M  \\
      \method(only $\mathcal{L}_{pred}$) & $\mathbf{0.2780}$   &  \underline{$0.3793$}  &  $\mathbf{0.4139}$  &  \underline{$0.4079$}  & $2.69$M  & $\mathbf{0.3401}$   &  $\mathbf{0.4344}$    & $\mathbf{0.4304}$    & $\mathbf{0.4318}$       & $3.59$M  \\
      \method  w/o. $\mathcal{L}_{distinct, alpha}$ & $0.2693$ & $0.3684$ & $0.3998$ & $0.3970$              & $2.69$M  & \underline{$0.3353$}     &  \underline{$0.4272$}    &  \underline{$0.4252$}      &   $0.4255$     & $3.59$M  \\
      \method w/o. MP & $0.2648$   &  $0.3650$  &  $0.4021$  &  $0.3960$  & $2.45$M  &  $0.3221$      &  $0.4172$   &  $0.4093$    &  $0.4131$      & $3.48$M  \\
      \method    & \underline{$0.2775^{*}$}   &  $\mathbf{0.3831}^{*}$  &  \underline{$0.4135$}  &  $\mathbf{0.4088}^{*}$  & $2.69$M  & $0.3344^{*}$   &  $0.4270$    & $0.4236^{*}$    & $0.4239$       & $3.59$M  \\ \bottomrule
    \end{tabular}%
  }
  \begin{tablenotes}
    \item a. Bolded values highlight the best performance, while underlined values denote the second-best.
    \item b. \method($K=1$) refers to the \method variant extracting only one temporal phenotype per patient.
    \item c. \method(only $\mathcal{L}_{pred}$) indicates \method trained without objectives related to interpretability.
    \item d. \method w/o. MP represents the \method variant without message passing on patient hypergraphs.
    \item e. An * indicates that the superiority of \method over the best-performing baseline is statistically significant.
  \end{tablenotes}
  \end{threeparttable}
}
\end{table*}

\subsection{Robustness Against False Negatives}
To quantitatively evaluate whether \method can effectively deal with incomplete patient EHRs, we randomly mask out $25\%$ and $75\%$ of input diagnoses in the test data to simulate varying levels of data incompleteness and assess the impact on \method's performance relative to three top-performing baselines. From \tableref{tab:robustness}, we observe that \method consistently surpasses all baselines across all metrics, affirming its robustness against false negative diagnoses. Notably, \method exhibits the least performance degradation, verifying its efficacy in contexts with incomplete data. Additionally, \method's phenotypes, despite their conciseness, recovered an average of $4.83\%$, $0.74\%$, $10.14\%$, and $1.67\%$ of the masked diagnoses across the four different settings, demonstrating \method's adeptness at handling data incompleteness. The inclusion of $\mathcal{L}_{distinct}$, which encourages the explanations to be concise, results in false negatives that are not critical in predicting future diagnoses being excluded from the extracted phenotypes, leading to a modest recovery rate. This also shows that with $\mathcal{L}_{distinct}$, the likelihood of \method adding true negatives is low.

\begin{table*}[ht]
\floatconts
  {tab:robustness}
  {\caption{Robustness against False Negatives under Four Settings}}
  {
  \begin{threeparttable}
\resizebox{\textwidth}{!}{
    \begin{tabular}{c|cccccccc}
      \hline
      Model  & Recall@$20$       & nDCG@$20$       & Recall@$20$        & nDCG@$20$        & Recall@$20$        & nDCG@$20$        & Recall@$20$        & nDCG@$20$        \\ \hline
      \multicolumn{1}{l|}{} &
      \multicolumn{2}{c}{MIMIC-III, masking $25\%$} &
      \multicolumn{2}{c}{MIMIC-III, masking $75\%$} &
      \multicolumn{2}{c}{MIMIC-IV, masking $25\%$} &
      \multicolumn{2}{c}{MIMIC-IV, masking $75\%$} \\ \hline
      CGL    & $0.3448$ ($-6.28\%$) & $0.3792$ ($-5.73\%$) & $0.2649$ ($-28.01\%$) & $0.2882$ ($-28.33\%$) & $0.3774$ ($-10.09\%$) & $0.3644$ ($-12.83\%$) & $0.2168$ ($-48.35\%$) & $0.2003$ ($-52.08\%$) \\
      T-LSTM & $0.3503$ ($-6.20\%$) & $0.3828$ ($-5.61\%$) & $0.2715$ ($-27.31\%$) & $0.2934$ ($-27.66\%$) & $0.3919$ ($-8.18\%$)  & $0.3943$ ($-6.82\%$)  & $0.3165$ ($-25.85\%$) & $0.2342$ ($-44.67\%$) \\
      AdaCare & $0.3443$ ($-6.18\%$) & $0.3740$ ($-4.95\%$) & $0.2890$ ($-21.25\%$) & $0.3173$ ($-19.37\%$) & $0.3816$ ($-10.39\%$) & $0.3659$ ($-13.31\%$) & $0.1749$ ($-58.94\%$) & $0.1489$ ($-64.71\%$) \\
      \method    & $\mathbf{0.3604}^{*}$ ($\mathbf{-5.92\%}$) & $\mathbf{0.3941}^{*}$ ($\mathbf{-3.58\%}$) & $\mathbf{0.3020}^{*}$ ($\mathbf{-21.17\%}$) & $\mathbf{0.3298}^{*}$ ($\mathbf{-19.32\%}$) & $\mathbf{0.3945}^{*}$ ($\mathbf{-7.61\%}$)  & $\mathbf{0.4150}^{*}$ ($\mathbf{-2.10\%}$)  & $\mathbf{0.3213}^{*}$ ($\mathbf{-24.75\%}$) & $\mathbf{0.3293}^{*}$ ($\mathbf{-22.31\%}$) \\ \hline
    \end{tabular}
  }
  \begin{tablenotes}
    \item a. The percentages reflect the relative performance drop from the setting with no past diagnoses masked.
  \end{tablenotes}
  \end{threeparttable}
}
\end{table*}

\subsection{Evaluation of Model Explanations}
\tableref{tab:expl_results} highlights the superior quality of explanations provided by \method compared to other self-explaining diagnosis prediction models. Notably, while all baseline models have Faithfulness scores below $0.5$, suggesting a weak to medium correlation between generated weights and prediction changes upon removal of the explanation units, \method showcases a much stronger correlation. Since RETAIN, AdaCare, and Timeline offer explanations by assigning a weight to every diagnosis, their Complexity score equals to the average number of historical diagnoses per patient in MIMIC-III. Dipole assigns visit-level attention scores, making the generated explanations too coarse to be eligible for calculating Complexity. Although \method provides explanations by extracting multiple phenotypes (subgraphs of the patient hypergraph) and introduces false negative disease-visit pairs before extraction, the total number of diagnoses in all phenotypes is, on average, lower than number of historical diagnoses the patient has in the record. This indicates that \method's phenotype extraction module effectively filters out irrelevant noise, yielding concise explanations.

A peculiar observation is the superior quality of explanations by \method w/o. $\mathcal{L}_{alpha}$, based on the three metrics. We looked into the explanations it offered, and found that for most of the patients, the generated $\boldsymbol{\alpha}$ were $\{1.0, 0.0, 0.0, 0.0, 0.0\}$, and only one of the five temporal phenotypes was non-empty. Based on how Faithfulness and Distinctness are calculated, it is reasonable for \method w/o. $\mathcal{L}_{alpha}$ to achieve outstanding results in these two metrics. \method w/o. $\mathcal{L}_{fidelity, alpha}$ shows a similar behavior. Thus, we can understand the importance of $\mathcal{L}_{alpha}$ in preventing the attention scores from being too extreme.

Comparing \method with \method w/o. $\mathcal{L}_{fidelity}$, \method w/o. $\mathcal{L}_{distinct}$ with \method w/o. $\mathcal{L}_{fidelity, distinct}$, \method w/o. $\mathcal{L}_{alpha}$ with \method w/o. $\mathcal{L}_{fidelity, alpha}$, and \method w/o. $\mathcal{L}_{distinct, alpha}$ with \method(only $\mathcal{L}_{pred}$) reveals that the inclusion of $\mathcal{L}_{fidelity}$ consistently enhances Faithfulness, albeit modestly. Moreover, removing $\mathcal{L}_{distinct}$ leads to substantial degradation across all metrics. The uptick in Complexity and downturn in Distinctness underscore its significance in generating concise, non-overlapping explanations. The pronounced decline in Faithfulness when excluding $\mathcal{L}_{distinct}$ stems from near-identical phenotypes having identical weights, making the calculation of correlation coefficients highly unstable. In conclusion, results from \tableref{tab:expl_results} affirm the importance of all three interpretability objectives, with $\mathcal{L}_{distinct}$ and $\mathcal{L}_{alpha}$ being particularly crucial.

\begin{table}[hbtp]
\floatconts
  {tab:expl_results}
  {\caption{Evaluation of Explanation Quality}}
  {
  \begin{threeparttable}
  \resizebox{\columnwidth}{!}{
    \begin{tabular}{@{}c|ccc@{}}
      \toprule
      Model                   & Faithfulness & Complexity & Distinctness \\ \midrule
                        & \multicolumn{3}{c}{MIMIC-III}        \\ \midrule
      RETAIN                  & $0.1867$       & $17.6564$     & --      \\
      Dipole                  & $0.3028$       & --          & --      \\
      AdaCare                 & $0.2084$       & $17.6564$     & --      \\
      Timeline                & $0.2499$       & $17.6564$    & --      \\ \midrule
      \method w/o. $\mathcal{L}_{fidelity}$             & $0.5206$     & $11.1233$   &  $0.7470$       \\
      \method w/o. $\mathcal{L}_{distinct}$          & $0.1139$     & $97.9960$   &  $0.2398$       \\
      \method w/o. $\mathcal{L}_{alpha}$             & $\mathbf{0.7381}$     & \underline{$5.0899$}    &  $\mathbf{1.0000}$       \\
      \method w/o. $\mathcal{L}_{fidelity, distinct}$ & $0.0789$     & $98.1879$   &  $0.2420$       \\
      \method w/o. $\mathcal{L}_{fidelity, alpha}$    & \underline{$0.7269$}     &   $\mathbf{5.0155}$          &   $\mathbf{1.0000}$      \\
      \method w/o. $\mathcal{L}_{distinct, alpha}$ & $0.0834$     & $103.2148$  &  $0.2295$       \\
      \method(only $\mathcal{L}_{pred}$)               & $0.0801$     & $103.6230$  &  $0.2293$       \\
      \method                 & $0.5915$     & $10.1839$   &  \underline{$0.7749$}       \\ \bottomrule
    \end{tabular}%
  }
  \begin{tablenotes}
    \item a. ``--'' means unavailable.
  \end{tablenotes}
  \end{threeparttable}}
\end{table}

\subsection{Case Study}
To showcase \method's capabilities in accommodating interventions from domain experts, we conduct a case study on a patient with three historical visits. \figureref{fig:case_study} illustrates this case study. \method accurately predicted $7$ out of $9$ diagnoses for the next visit. In particular, \method anticipated the onset of a urinary tract infection, despite the lack of information on the timing of the next visit, and in the absence of any previous records explicitly indicating urinary tract problems for the patient. Additionally, \method identified acidosis, a condition that was not recorded in the patient's last three visits. Thus, \method demonstrated capabilities beyond merely replicating past diagnoses. Interestingly, \method's prediction listed diabetes, which was not confirmed by the ground-truth label. However, considering the patient's recent diagnosis of cirrhosis of liver—a condition affecting insulin sensitivity—and the chronic nature of diabetes mellitus, this prediction still holds merit.

\begin{figure*}[ht]
\floatconts
  {fig:case_study}
  {\caption{An illustration of how \method extracts five temporal phenotypes from the EHR of a 53-year-old female patient and how a clinician can refine the prediction by directly adjusting the generated phenotypes. Underlined text highlights human modifications, while check marks indicate correct predictions. The \textcolor{red}{red} numbers indicate the importance weights.}}
  {\includegraphics[width=\textwidth]{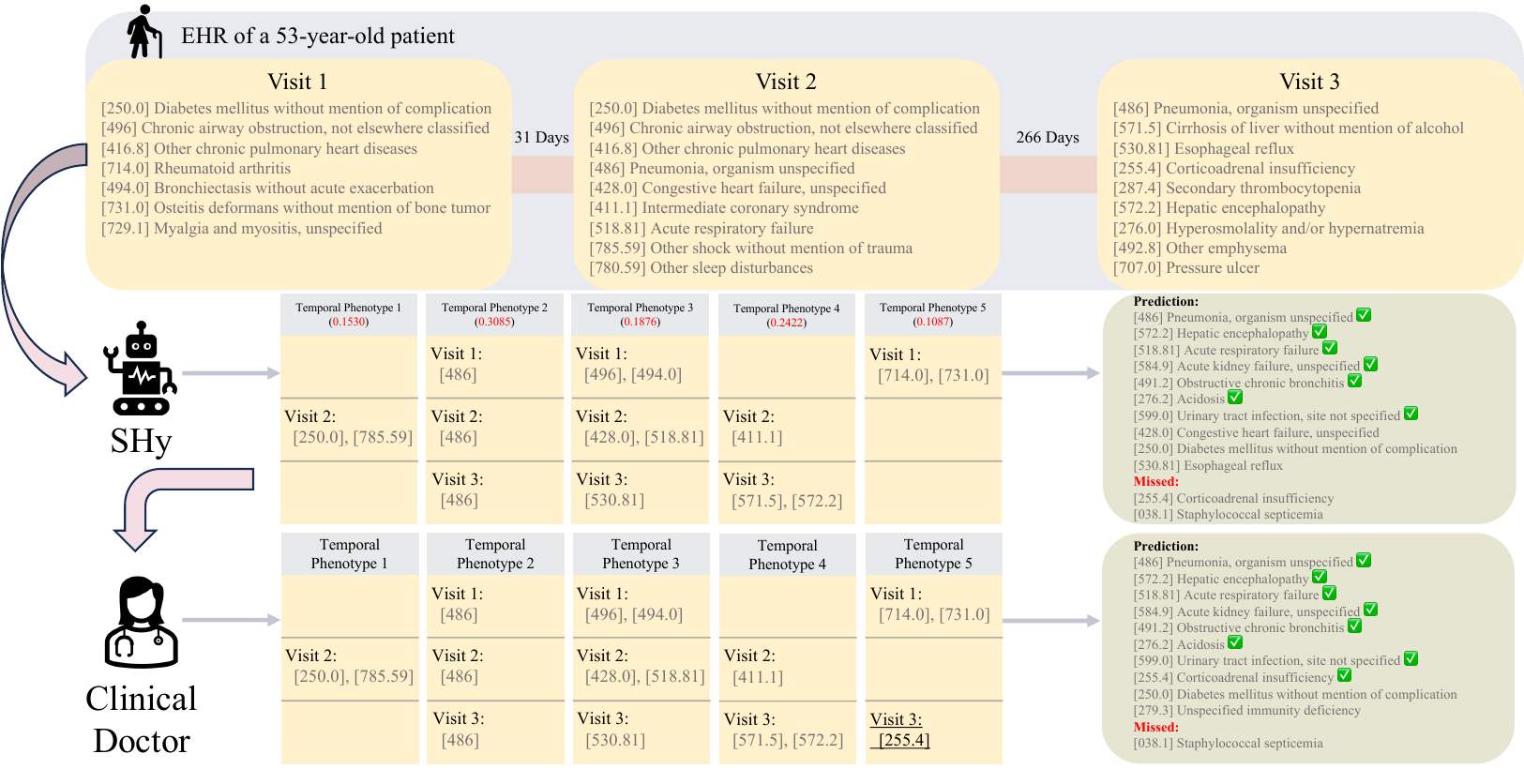}}
\end{figure*}

\method offered compelling rationales for its predictions. \textbf{Each extracted phenotype reflected a unique facet of the patient’s health status.} Phenotype $1$ indicated that the patient might have experienced endocrine system abnormalities, possibly even severe hypoglycemia, due to diabetes; Phenotype $2$, which consistently highlighted pneumonia at all visits, symbolized worsening respiratory health and a compromised immune system; Phenotype $3$ underlined pneumonia complications and congestive heart failure, suggesting a high probability of hypoxia; Phenotype $4$ indicated significant liver impairment and resulting coronary artery lesions; Phenotype $5$ implied that the abnormal immune response was detrimentally affecting the bones and joints. \textbf{Every phenotype contributed to the prediction result.} Pneumonia, chronic obstructive bronchitis and acute respiratory failure could be inferred from Phenotype $2$; urinary tract infection, closely associated with diabetes and acute kidney failure, with hypoxia as the main driving force, could be anticipated based on Phenotype $1$ and $3$; hepatic encephalopathy could be predicted through Phenotype $4$; Phenotype $5$, reflecting the patient's waning immune system, played a suggestive role in predicting most lesions. \textbf{The weights assigned to phenotypes were justifiable.} Phenotype $2$, which received the highest weight, was the most critical because the patient had diagnoses related to the respiratory system across all past and subsequent visits, and a consistent diagnosis of pneumonia suggested a weakening immune system and a critical health condition; Phenotype $5$, assigned the lowest weight, was less relevant compared to the other phenotypes, as the patient did not receive diagnoses similar to those within this phenotype during the last two visits. More analysis on the case study can be found in \appendixref{apd:fifth}.

\section{Conclusions}
In this study, we introduce \method, a self-explaining model for diagnosis prediction. It represents each patient's longitudinal EHR as a hypergraph and employs message passing to derive personalized disease embeddings. With these embeddings, \method offers tailored explanations by extracting temporal phenotypes upon which predictions are based. With a novel combination of objectives, we ensure that these temporal phenotypes are concise, faithful, distinct, and easy to be edited by domain experts. Quantitative evaluations highlight \method's competitive predictive performance and superior explanatory capabilities, advocating its potential as an AI assistant. Future endeavors include employing timestamps in the prediction process and devising more robust metrics for evaluating the explanation quality of diagnosis prediction models.


\bibliography{chil-sample}

\begin{thebibliography}{78}
\providecommand{\natexlab}[1]{#1}
\providecommand{\url}[1]{\texttt{#1}}
\expandafter\ifx\csname urlstyle\endcsname\relax
  \providecommand{\doi}[1]{doi: #1}\else
  \providecommand{\doi}{doi: \begingroup \urlstyle{rm}\Url}\fi

\bibitem[Alvarez~Melis and Jaakkola(2018)]{using:4}
David Alvarez~Melis and Tommi Jaakkola.
\newblock Towards robust interpretability with self-explaining neural networks.
\newblock \emph{Advances in neural information processing systems}, 31, 2018.

\bibitem[Arya et~al.(2020)Arya, Gupta, Rudinac, and Worring]{hyper:11}
Devanshu Arya, Deepak~K Gupta, Stevan Rudinac, and Marcel Worring.
\newblock Hypersage: Generalizing inductive representation learning on hypergraphs, 2020.

\bibitem[Bai et~al.(2021)Bai, Zhang, and Torr]{hyper:8}
Song Bai, Feihu Zhang, and Philip~HS Torr.
\newblock Hypergraph convolution and hypergraph attention.
\newblock \emph{Pattern Recognition}, 110:\penalty0 107637, 2021.

\bibitem[Bai et~al.(2018)Bai, Zhang, Egleston, and Vucetic]{baseline:12}
Tian Bai, Shanshan Zhang, Brian~L Egleston, and Slobodan Vucetic.
\newblock Interpretable representation learning for healthcare via capturing disease progression through time.
\newblock In \emph{Proceedings of the 24th ACM SIGKDD international conference on knowledge discovery \& data mining}, pages 43--51, 2018.

\bibitem[Baytas et~al.(2017)Baytas, Xiao, Zhang, Wang, Jain, and Zhou]{baseline:11}
Inci~M Baytas, Cao Xiao, Xi~Zhang, Fei Wang, Anil~K Jain, and Jiayu Zhou.
\newblock Patient subtyping via time-aware lstm networks.
\newblock In \emph{Proceedings of the 23rd ACM SIGKDD international conference on knowledge discovery and data mining}, 2017.

\bibitem[Cai et~al.(2022{\natexlab{a}})Cai, Song, Sun, Zhang, Hong, and Li]{using:3}
Derun Cai, Moxian Song, Chenxi Sun, Baofeng Zhang, Shenda Hong, and Hongyan Li.
\newblock Hypergraph structure learning for hypergraph neural networks.
\newblock In \emph{Proceedings of the Thirty-First International Joint Conference on Artificial Intelligence, IJCAI-22}, pages 1923--1929, 2022{\natexlab{a}}.

\bibitem[Cai et~al.(2022{\natexlab{b}})Cai, Sun, Song, Zhang, Hong, and Li]{hyper:2}
Derun Cai, Chenxi Sun, Moxian Song, Baofeng Zhang, Shenda Hong, and Hongyan Li.
\newblock Hypergraph contrastive learning for electronic health records.
\newblock In \emph{Proceedings of the 2022 SIAM International Conference on Data Mining (SDM)}, pages 127--135. SIAM, 2022{\natexlab{b}}.

\bibitem[Cai et~al.(2022{\natexlab{c}})Cai, Zheng, Ooi, Wang, and Yao]{ehr:7}
Qingpeng Cai, Kaiping Zheng, Beng~Chin Ooi, Wei Wang, and Chang Yao.
\newblock Elda: Learning explicit dual-interactions for healthcare analytics.
\newblock In \emph{2022 IEEE 38th International Conference on Data Engineering (ICDE)}, pages 393--406. IEEE, 2022{\natexlab{c}}.

\bibitem[Che et~al.(2015)Che, Kale, Li, Bahadori, and Liu]{ehr:16}
Zhengping Che, David Kale, Wenzhe Li, Mohammad~Taha Bahadori, and Yan Liu.
\newblock Deep computational phenotyping.
\newblock In \emph{Proceedings of the 21th ACM SIGKDD International Conference on Knowledge Discovery and Data Mining}, pages 507--516, 2015.

\bibitem[Chien et~al.(2022)Chien, Pan, Peng, and Milenkovic]{hyper:4}
Eli Chien, Chao Pan, Jianhao Peng, and Olgica Milenkovic.
\newblock You are allset: A multiset function framework for hypergraph neural networks.
\newblock In \emph{International Conference on Learning Representations}, 2022.

\bibitem[Choi et~al.(2016{\natexlab{a}})Choi, Bahadori, Schuetz, Stewart, and Sun]{baseline:1}
Edward Choi, Mohammad~Taha Bahadori, Andy Schuetz, Walter~F Stewart, and Jimeng Sun.
\newblock Doctor ai: Predicting clinical events via recurrent neural networks.
\newblock In \emph{Machine learning for healthcare conference}, pages 301--318. PMLR, 2016{\natexlab{a}}.

\bibitem[Choi et~al.(2016{\natexlab{b}})Choi, Bahadori, Searles, Coffey, Thompson, Bost, Tejedor-Sojo, and Sun]{baseline:16}
Edward Choi, Mohammad~Taha Bahadori, Elizabeth Searles, Catherine Coffey, Michael Thompson, James Bost, Javier Tejedor-Sojo, and Jimeng Sun.
\newblock Multi-layer representation learning for medical concepts.
\newblock In \emph{proceedings of the 22nd ACM SIGKDD international conference on knowledge discovery and data mining}, pages 1495--1504, 2016{\natexlab{b}}.

\bibitem[Choi et~al.(2016{\natexlab{c}})Choi, Bahadori, Sun, Kulas, Schuetz, and Stewart]{baseline:2}
Edward Choi, Mohammad~Taha Bahadori, Jimeng Sun, Joshua Kulas, Andy Schuetz, and Walter Stewart.
\newblock Retain: An interpretable predictive model for healthcare using reverse time attention mechanism.
\newblock \emph{Advances in neural information processing systems}, 29, 2016{\natexlab{c}}.

\bibitem[Choi et~al.(2017)Choi, Bahadori, Song, Stewart, and Sun]{baseline:6}
Edward Choi, Mohammad~Taha Bahadori, Le~Song, Walter~F Stewart, and Jimeng Sun.
\newblock Gram: graph-based attention model for healthcare representation learning.
\newblock In \emph{Proceedings of the 23rd ACM SIGKDD international conference on knowledge discovery and data mining}, pages 787--795, 2017.

\bibitem[Choi et~al.(2018)Choi, Xiao, Stewart, and Sun]{ehr:3}
Edward Choi, Cao Xiao, Walter Stewart, and Jimeng Sun.
\newblock Mime: Multilevel medical embedding of electronic health records for predictive healthcare.
\newblock \emph{Advances in neural information processing systems}, 31, 2018.

\bibitem[Cui et~al.(2022)Cui, Luo, Ye, Wang, Wang, and Ma]{ehr:5}
Suhan Cui, Junyu Luo, Muchao Ye, Jiaqi Wang, Ting Wang, and Fenglong Ma.
\newblock Medskim: Denoised health risk prediction via skimming medical claims data.
\newblock In \emph{2022 IEEE International Conference on Data Mining (ICDM)}, pages 81--90. IEEE, 2022.

\bibitem[Dong et~al.(2020)Dong, Sawin, and Bengio]{hyper:13}
Yihe Dong, Will Sawin, and Yoshua Bengio.
\newblock Hnhn: Hypergraph networks with hyperedge neurons, 2020.

\bibitem[Du et~al.(2019)Du, Liu, and Hu]{ii:1}
Mengnan Du, Ninghao Liu, and Xia Hu.
\newblock Techniques for interpretable machine learning.
\newblock \emph{Communications of the ACM}, 63\penalty0 (1):\penalty0 68--77, 2019.

\bibitem[Feng et~al.(2019)Feng, You, Zhang, Ji, and Gao]{hyper:5}
Yifan Feng, Haoxuan You, Zizhao Zhang, Rongrong Ji, and Yue Gao.
\newblock Hypergraph neural networks.
\newblock In \emph{Proceedings of the AAAI conference on artificial intelligence}, pages 3558--3565, 2019.

\bibitem[Gao et~al.(2020)Gao, Xiao, Wang, Tang, Glass, and Sun]{baseline:4}
Junyi Gao, Cao Xiao, Yasha Wang, Wen Tang, Lucas~M Glass, and Jimeng Sun.
\newblock Stagenet: Stage-aware neural networks for health risk prediction.
\newblock In \emph{Proceedings of The Web Conference 2020}, pages 530--540, 2020.

\bibitem[Gao et~al.(2022)Gao, Feng, Ji, and Ji]{hyper:7}
Yue Gao, Yifan Feng, Shuyi Ji, and Rongrong Ji.
\newblock Hgnn+: General hypergraph neural networks.
\newblock \emph{IEEE Transactions on Pattern Analysis and Machine Intelligence}, 45\penalty0 (3):\penalty0 3181--3199, 2022.

\bibitem[Gautam et~al.(2022)Gautam, Boubekki, Hansen, Salahuddin, Jenssen, H{\"o}hne, and Kampffmeyer]{ii:5}
Srishti Gautam, Ahcene Boubekki, Stine Hansen, Suaiba Salahuddin, Robert Jenssen, Marina H{\"o}hne, and Michael Kampffmeyer.
\newblock Protovae: A trustworthy self-explainable prototypical variational model.
\newblock \emph{Advances in Neural Information Processing Systems}, 35:\penalty0 17940--17952, 2022.

\bibitem[Goldberger et~al.(2000)Goldberger, Amaral, Glass, Hausdorff, Ivanov, Mark, Mietus, Moody, Peng, and Stanley]{other:3}
Ary~L Goldberger, Luis~AN Amaral, Leon Glass, Jeffrey~M Hausdorff, Plamen~Ch Ivanov, Roger~G Mark, Joseph~E Mietus, George~B Moody, Chung-Kang Peng, and H~Eugene Stanley.
\newblock Physiobank, physiotoolkit, and physionet: components of a new research resource for complex physiologic signals.
\newblock \emph{circulation}, pages e215--e220, 2000.

\bibitem[Huang and Yang(2021)]{using:8}
Jing Huang and Jie Yang.
\newblock Unignn: a unified framework for graph and hypergraph neural networks.
\newblock In \emph{Proceedings of the Thirty-First International Joint Conference on Artificial Intelligence, IJCAI-21}, 2021.

\bibitem[Jain and Wallace(2019)]{ii:11}
Sarthak Jain and Byron~C Wallace.
\newblock Attention is not explanation.
\newblock In \emph{Proceedings of NAACL-HLT}, 2019.

\bibitem[Jang et~al.(2017)Jang, Gu, and Poole]{using:7}
Eric Jang, Shixiang Gu, and Ben Poole.
\newblock Categorical reparameterization with gumbel-softmax.
\newblock In \emph{International Conference on Learning Representations}, 2017.

\bibitem[Jo et~al.(2021)Jo, Baek, Lee, Kim, Kang, and Hwang]{hyper:12}
Jaehyeong Jo, Jinheon Baek, Seul Lee, Dongki Kim, Minki Kang, and Sung~Ju Hwang.
\newblock Edge representation learning with hypergraphs.
\newblock \emph{Advances in Neural Information Processing Systems}, 34:\penalty0 7534--7546, 2021.

\bibitem[Johnson et~al.(2016{\natexlab{a}})Johnson, Pollard, and Mark]{johnson2016physionet}
Alistair E.~W. Johnson, Tom~J. Pollard, and Roger~G. Mark.
\newblock {MIMIC-III} clinical database (version 1.4), 2016{\natexlab{a}}.

\bibitem[Johnson et~al.(2016{\natexlab{b}})Johnson, Pollard, Shen, Lehman, Feng, Ghassemi, Moody, Szolovits, Celi, and Mark]{johnson2016mimic}
Alistair E.~W. Johnson, Tom~J. Pollard, Lu~Shen, Li-wei~H. Lehman, Mengling Feng, Mohammad Ghassemi, Benjamin Moody, Peter Szolovits, Leo~Anthony Celi, and Roger~G. Mark.
\newblock {MIMIC-III}, a freely accessible critical care database.
\newblock \emph{Scientific Data}, 3\penalty0 (160035), 2016{\natexlab{b}}.
\newblock \doi{https://doi.org/10.1038/sdata.2016.35}.

\bibitem[Johnson et~al.(2023)Johnson, Bulgarelli, Shen, Gayles, Shammout, Horng, Pollard, Hao, Moody, Gow, et~al.]{johnson2023mimic}
Alistair~EW Johnson, Lucas Bulgarelli, Lu~Shen, Alvin Gayles, Ayad Shammout, Steven Horng, Tom~J Pollard, Sicheng Hao, Benjamin Moody, Brian Gow, et~al.
\newblock Mimic-iv, a freely accessible electronic health record dataset.
\newblock \emph{Scientific data}, 10\penalty0 (1):\penalty0 1, 2023.

\bibitem[Kipf and Welling(2017)]{other:1}
Thomas~N Kipf and Max Welling.
\newblock Semi-supervised classification with graph convolutional networks.
\newblock In \emph{International Conference on Learning Representations}, 2017.

\bibitem[Koh et~al.(2020)Koh, Nguyen, Tang, Mussmann, Pierson, Kim, and Liang]{using:6}
Pang~Wei Koh, Thao Nguyen, Yew~Siang Tang, Stephen Mussmann, Emma Pierson, Been Kim, and Percy Liang.
\newblock Concept bottleneck models.
\newblock In \emph{International conference on machine learning}, pages 5338--5348. PMLR, 2020.

\bibitem[Li et~al.(2018)Li, Liu, Chen, and Rudin]{using:5}
Oscar Li, Hao Liu, Chaofan Chen, and Cynthia Rudin.
\newblock Deep learning for case-based reasoning through prototypes: A neural network that explains its predictions.
\newblock In \emph{Proceedings of the AAAI Conference on Artificial Intelligence}, 2018.

\bibitem[Li et~al.(2023)Li, Mamouei, Salimi-Khorshidi, Rao, Hassaine, Canoy, Lukasiewicz, and Rahimi]{ehr:18}
Yikuan Li, Mohammad Mamouei, Gholamreza Salimi-Khorshidi, Shishir Rao, Abdelaali Hassaine, Dexter Canoy, Thomas Lukasiewicz, and Kazem Rahimi.
\newblock Hi-behrt: Hierarchical transformer-based model for accurate prediction of clinical events using multimodal longitudinal electronic health records.
\newblock \emph{IEEE Journal of Biomedical and Health Informatics}, November 2023.
\newblock URL \url{https://doi.org/10.1109/JBHI.2022.3224727}.

\bibitem[Liu et~al.(2015)Liu, Wang, Hu, and Xiong]{using:1}
Chuanren Liu, Fei Wang, Jianying Hu, and Hui Xiong.
\newblock Temporal phenotyping from longitudinal electronic health records: A graph based framework.
\newblock In \emph{Proceedings of the 21th ACM SIGKDD international conference on knowledge discovery and data mining}, pages 705--714, 2015.

\bibitem[Lu et~al.(2021{\natexlab{a}})Lu, Reddy, Chakraborty, Kleinberg, and Ning]{baseline:7}
Chang Lu, Chandan~K Reddy, Prithwish Chakraborty, Samantha Kleinberg, and Yue Ning.
\newblock Collaborative graph learning with auxiliary text for temporal event prediction in healthcare.
\newblock In \emph{Proceedings of the Thirty-First International Joint Conference on Artificial Intelligence, IJCAI-21}, 2021{\natexlab{a}}.

\bibitem[Lu et~al.(2021{\natexlab{b}})Lu, Reddy, and Ning]{baseline:14}
Chang Lu, Chandan~K Reddy, and Yue Ning.
\newblock Self-supervised graph learning with hyperbolic embedding for temporal health event prediction.
\newblock \emph{IEEE Transactions on Cybernetics}, 2021{\natexlab{b}}.

\bibitem[Lu et~al.(2022)Lu, Han, and Ning]{baseline:10}
Chang Lu, Tian Han, and Yue Ning.
\newblock Context-aware health event prediction via transition functions on dynamic disease graphs.
\newblock In \emph{Proceedings of the AAAI Conference on Artificial Intelligence}, volume~36, pages 4567--4574, 2022.

\bibitem[Luo et~al.(2020)Luo, Ye, Xiao, and Ma]{baseline:5}
Junyu Luo, Muchao Ye, Cao Xiao, and Fenglong Ma.
\newblock Hitanet: Hierarchical time-aware attention networks for risk prediction on electronic health records.
\newblock In \emph{Proceedings of the 26th ACM SIGKDD International Conference on Knowledge Discovery \& Data Mining}, pages 647--656, 2020.

\bibitem[Ma et~al.(2017)Ma, Chitta, Zhou, You, Sun, and Gao]{baseline:3}
Fenglong Ma, Radha Chitta, Jing Zhou, Quanzeng You, Tong Sun, and Jing Gao.
\newblock Dipole: Diagnosis prediction in healthcare via attention-based bidirectional recurrent neural networks.
\newblock In \emph{Proceedings of the 23rd ACM SIGKDD international conference on knowledge discovery and data mining}, pages 1903--1911, 2017.

\bibitem[Ma et~al.(2018{\natexlab{a}})Ma, Gao, Suo, You, Zhou, and Zhang]{ehr:21}
Fenglong Ma, Jing Gao, Qiuling Suo, Quanzeng You, Jing Zhou, and Aidong Zhang.
\newblock Risk prediction on electronic health records with prior medical knowledge.
\newblock In \emph{Proceedings of the 24th ACM SIGKDD International Conference on Knowledge Discovery \& Data Mining}, 2018{\natexlab{a}}.

\bibitem[Ma et~al.(2018{\natexlab{b}})Ma, You, Xiao, Chitta, Zhou, and Gao]{ehr:1}
Fenglong Ma, Quanzeng You, Houping Xiao, Radha Chitta, Jing Zhou, and Jing Gao.
\newblock Kame: Knowledge-based attention model for diagnosis prediction in healthcare.
\newblock In \emph{Proceedings of the 27th ACM International Conference on Information and Knowledge Management}, pages 743--752, 2018{\natexlab{b}}.

\bibitem[Ma et~al.(2020{\natexlab{a}})Ma, Gao, Wang, Zhang, Wang, Ruan, Tang, Gao, and Ma]{baseline:8}
Liantao Ma, Junyi Gao, Yasha Wang, Chaohe Zhang, Jiangtao Wang, Wenjie Ruan, Wen Tang, Xin Gao, and Xinyu Ma.
\newblock Adacare: Explainable clinical health status representation learning via scale-adaptive feature extraction and recalibration.
\newblock In \emph{Proceedings of the AAAI Conference on Artificial Intelligence}, pages 825--832, 2020{\natexlab{a}}.

\bibitem[Ma et~al.(2020{\natexlab{b}})Ma, Zhang, Wang, Ruan, Wang, Tang, Ma, Gao, and Gao]{baseline:9}
Liantao Ma, Chaohe Zhang, Yasha Wang, Wenjie Ruan, Jiangtao Wang, Wen Tang, Xinyu Ma, Xin Gao, and Junyi Gao.
\newblock Concare: Personalized clinical feature embedding via capturing the healthcare context.
\newblock In \emph{Proceedings of the AAAI Conference on Artificial Intelligence}, 2020{\natexlab{b}}.

\bibitem[Maddison et~al.(2017)Maddison, Mnih, and Teh]{using:11}
Chris~J Maddison, Andriy Mnih, and Yee~Whye Teh.
\newblock The concrete distribution: A continuous relaxation of discrete random variables.
\newblock In \emph{International Conference on Learning Representations}, 2017.

\bibitem[Meng et~al.(2022)Meng, Trinh, Xu, Enouen, and Liu]{other:10}
Chuizheng Meng, Loc Trinh, Nan Xu, James Enouen, and Yan Liu.
\newblock Interpretability and fairness evaluation of deep learning models on mimic-iv dataset.
\newblock \emph{Scientific Reports}, 2022.

\bibitem[Mohankumar et~al.(2020)Mohankumar, Nema, Narasimhan, Khapra, Srinivasan, and Ravindran]{ii:6}
Akash~Kumar Mohankumar, Preksha Nema, Sharan Narasimhan, Mitesh~M Khapra, Balaji~Vasan Srinivasan, and Balaraman Ravindran.
\newblock Towards transparent and explainable attention models.
\newblock In \emph{Proceedings of the 58th Annual Meeting of the Association for Computational Linguistics}, pages 4206--4216, 2020.

\bibitem[Payrovnaziri et~al.(2020)Payrovnaziri, Chen, Rengifo-Moreno, Miller, Bian, Chen, Liu, and He]{ehr:17}
Seyedeh~Neelufar Payrovnaziri, Zhaoyi Chen, Pablo Rengifo-Moreno, Tim Miller, Jiang Bian, Jonathan~H Chen, Xiuwen Liu, and Zhe He.
\newblock Explainable artificial intelligence models using real-world electronic health record data: a systematic scoping review.
\newblock \emph{Journal of the American Medical Informatics Association}, 27\penalty0 (7):\penalty0 1173--1185, 2020.

\bibitem[Peng et~al.(2021{\natexlab{a}})Peng, Long, Shen, Wang, and Jiang]{baseline:13}
Xueping Peng, Guodong Long, Tao Shen, Sen Wang, and Jing Jiang.
\newblock Sequential diagnosis prediction with transformer and ontological representation.
\newblock In \emph{2021 IEEE International Conference on Data Mining (ICDM)}, pages 489--498. IEEE, 2021{\natexlab{a}}.

\bibitem[Peng et~al.(2021{\natexlab{b}})Peng, Long, Wang, Jiang, Clarke, Schlegel, and Zhang]{baseline:15}
Xueping Peng, Guodong Long, Sen Wang, Jing Jiang, Allison Clarke, Clement Schlegel, and Chengqi Zhang.
\newblock Mipo: Mutual integration of patient journey and medical ontology for healthcare representation learning, 2021{\natexlab{b}}.

\bibitem[Qiao et~al.(2018)Qiao, Zhao, Xiao, Li, Qin, and Wang]{ehr:19}
Zhi Qiao, Shiwan Zhao, Cao Xiao, Xiang Li, Yong Qin, and Fei Wang.
\newblock Pairwise-ranking based collaborative recurrent neural networks for clinical event prediction.
\newblock In \emph{Proceedings of the twenty-seventh international joint conference on artificial intelligence}, 2018.

\bibitem[Qiao et~al.(2020)Qiao, Zhang, Wu, Ge, and Fan]{ehr:4}
Zhi Qiao, Zhen Zhang, Xian Wu, Shen Ge, and Wei Fan.
\newblock Mhm: Multi-modal clinical data based hierarchical multi-label diagnosis prediction.
\newblock In \emph{Proceedings of the 43rd International ACM SIGIR Conference on Research and Development in Information Retrieval}, pages 1841--1844, 2020.

\bibitem[Rasmy et~al.(2021)Rasmy, Xiang, Xie, Tao, and Zhi]{ehr:6}
Laila Rasmy, Yang Xiang, Ziqian Xie, Cui Tao, and Degui Zhi.
\newblock Med-bert: pretrained contextualized embeddings on large-scale structured electronic health records for disease prediction.
\newblock \emph{NPJ digital medicine}, 4\penalty0 (1):\penalty0 86, 2021.

\bibitem[Ren et~al.(2021)Ren, Wang, Zhao, and Wu]{ehr:20}
Houxing Ren, Jingyuan Wang, Wayne~Xin Zhao, and Ning Wu.
\newblock Rapt: Pre-training of time-aware transformer for learning robust healthcare representation.
\newblock In \emph{Proceedings of the 27th ACM SIGKDD conference on knowledge discovery \& data mining}, 2021.

\bibitem[Rudin(2019)]{ii:2}
Cynthia Rudin.
\newblock Stop explaining black box machine learning models for high stakes decisions and use interpretable models instead.
\newblock \emph{Nature machine intelligence}, 1\penalty0 (5):\penalty0 206--215, 2019.

\bibitem[Serrano and Smith(2019)]{ii:7}
Sofia Serrano and Noah~A Smith.
\newblock Is attention interpretable?
\newblock In \emph{Proceedings of the 57th Annual Meeting of the Association for Computational Linguistics}, pages 2931--2951, 2019.

\bibitem[Si et~al.(2021)Si, Du, Li, Jiang, Miller, Wang, Zheng, and Roberts]{ehr:11}
Yuqi Si, Jingcheng Du, Zhao Li, Xiaoqian Jiang, Timothy Miller, Fei Wang, W~Jim Zheng, and Kirk Roberts.
\newblock Deep representation learning of patient data from electronic health records (ehr): A systematic review.
\newblock \emph{Journal of biomedical informatics}, 115:\penalty0 103671, 2021.

\bibitem[Song et~al.(2018)Song, Rajan, Thiagarajan, and Spanias]{other:8}
Huan Song, Deepta Rajan, Jayaraman Thiagarajan, and Andreas Spanias.
\newblock Attend and diagnose: Clinical time series analysis using attention models.
\newblock In \emph{Proceedings of the AAAI conference on artificial intelligence}, 2018.

\bibitem[Song et~al.(2019)Song, Cheong, Yin, Cheung, Fung, and Poon]{baseline:17}
Lihong Song, Chin~Wang Cheong, Kejing Yin, William~K Cheung, Benjamin~CM Fung, and Jonathan Poon.
\newblock Medical concept embedding with multiple ontological representations.
\newblock In \emph{IJCAI}, volume~19, pages 4613--4619, 2019.

\bibitem[Tan et~al.(2022)Tan, Kong, Yu, Li, Chen, Zheng, Hertzberg, and Yang]{other:7}
Yanchao Tan, Chengjun Kong, Leisheng Yu, Pan Li, Chaochao Chen, Xiaolin Zheng, Vicki~S Hertzberg, and Carl Yang.
\newblock 4sdrug: Symptom-based set-to-set small and safe drug recommendation.
\newblock In \emph{Proceedings of the 28th ACM SIGKDD Conference on Knowledge Discovery and Data Mining}, 2022.

\bibitem[Tan et~al.(2023)Tan, Zhou, Yu, Liu, Chen, Ma, Hu, Hertzberg, and Yang]{tan2023enhancing}
Yanchao Tan, Zihao Zhou, Leisheng Yu, Weiming Liu, Chaochao Chen, Guofang Ma, Xiao Hu, Vicki~S Hertzberg, and Carl Yang.
\newblock Enhancing personalized healthcare via capturing disease severity, interaction, and progression.
\newblock In \emph{2023 IEEE International Conference on Data Mining (ICDM)}, 2023.

\bibitem[Tan et~al.(2024)Tan, Zhang, Zhou, Ma, Wang, Liu, Liao, Hertzberg, and Yang]{tan2024enhancing}
Yanchao Tan, Hengyu Zhang, Zihao Zhou, Guofang Ma, Fan Wang, Weiming Liu, Xinting Liao, Vicki~S Hertzberg, and Carl Yang.
\newblock Enhancing progressive diagnosis prediction in healthcare with continuous normalizing flows.
\newblock In \emph{Companion Proceedings of the ACM on Web Conference 2024}, 2024.

\bibitem[Tonekaboni et~al.(2019)Tonekaboni, Joshi, McCradden, and Goldenberg]{ehr:23}
Sana Tonekaboni, Shalmali Joshi, Melissa~D McCradden, and Anna Goldenberg.
\newblock What clinicians want: contextualizing explainable machine learning for clinical end use.
\newblock In \emph{Machine learning for healthcare conference}. PMLR, 2019.

\bibitem[Van~Smeden et~al.(2021)Van~Smeden, Reitsma, Riley, Collins, and Moons]{van2021clinical}
Maarten Van~Smeden, Johannes~B Reitsma, Richard~D Riley, Gary~S Collins, and Karel~GM Moons.
\newblock Clinical prediction models: diagnosis versus prognosis.
\newblock \emph{Journal of clinical epidemiology}, 132:\penalty0 142--145, 2021.

\bibitem[Wang and Wang(2021)]{ii:4}
Yipei Wang and Xiaoqian Wang.
\newblock Self-interpretable model with transformation equivariant interpretation.
\newblock \emph{Advances in Neural Information Processing Systems}, 34:\penalty0 2359--2372, 2021.

\bibitem[Wiegreffe and Pinter(2019)]{ii:10}
Sarah Wiegreffe and Yuval Pinter.
\newblock Attention is not not explanation.
\newblock In \emph{Proceedings of the 2019 Conference on Empirical Methods in Natural Language Processing and the 9th International Joint Conference on Natural Language Processing (EMNLP-IJCNLP)}, 2019.

\bibitem[Wu and Ng(2022)]{hyper:3}
Hanrui Wu and Michael~K Ng.
\newblock Hypergraph convolution on nodes-hyperedges network for semi-supervised node classification.
\newblock \emph{ACM Transactions on Knowledge Discovery from Data (TKDD)}, 16\penalty0 (4):\penalty0 1--19, 2022.

\bibitem[Xu et~al.(2018)Xu, Hu, Leskovec, and Jegelka]{other:5}
Keyulu Xu, Weihua Hu, Jure Leskovec, and Stefanie Jegelka.
\newblock How powerful are graph neural networks?
\newblock In \emph{International Conference on Learning Representations}, 2018.

\bibitem[Xu et~al.(2022)Xu, Yu, Zhang, Ali, Ho, and Yang]{hyper:1}
Ran Xu, Yue Yu, Chao Zhang, Mohammed~K Ali, Joyce~C Ho, and Carl Yang.
\newblock Counterfactual and factual reasoning over hypergraphs for interpretable clinical predictions on ehr.
\newblock In \emph{Machine Learning for Health}, pages 259--278. PMLR, 2022.

\bibitem[Xu et~al.(2023)Xu, Chu, Yang, Wang, Zou, Ding, Zhao, Wang, and Xie]{ehr:2}
Yongxin Xu, Xu~Chu, Kai Yang, Zhiyuan Wang, Peinie Zou, Hongxin Ding, Junfeng Zhao, Yasha Wang, and Bing Xie.
\newblock Seqcare: Sequential training with external medical knowledge graph for diagnosis prediction in healthcare data.
\newblock In \emph{Proceedings of the ACM Web Conference 2023}, pages 2819--2830, 2023.

\bibitem[Yadati et~al.(2019)Yadati, Nimishakavi, Yadav, Nitin, Louis, and Talukdar]{hyper:14}
Naganand Yadati, Madhav Nimishakavi, Prateek Yadav, Vikram Nitin, Anand Louis, and Partha Talukdar.
\newblock Hypergcn: A new method for training graph convolutional networks on hypergraphs.
\newblock \emph{Advances in neural information processing systems}, 32, 2019.

\bibitem[Zhang et~al.(2023)Zhang, Zheng, Zhou, and Lu]{ii:9}
Borui Zhang, Wenzhao Zheng, Jie Zhou, and Jiwen Lu.
\newblock Bort: Towards explainable neural networks with bounded orthogonal constraint.
\newblock In \emph{The Eleventh International Conference on Learning Representations}, 2023.

\bibitem[Zhang et~al.(2021)Zhang, Gao, Ma, Wang, Wang, and Tang]{ehr:22}
Chaohe Zhang, Xin Gao, Liantao Ma, Yasha Wang, Jiangtao Wang, and Wen Tang.
\newblock Grasp: generic framework for health status representation learning based on incorporating knowledge from similar patients.
\newblock In \emph{Proceedings of the AAAI conference on artificial intelligence}, 2021.

\bibitem[Zhang et~al.(2018)Zhang, Kowsari, Harrison, Lobo, and Barnes]{ehr:13}
Jinghe Zhang, Kamran Kowsari, James~H Harrison, Jennifer~M Lobo, and Laura~E Barnes.
\newblock Patient2vec: A personalized interpretable deep representation of the longitudinal electronic health record.
\newblock \emph{IEEE Access}, 6:\penalty0 65333--65346, 2018.

\bibitem[Zhang et~al.(2020{\natexlab{a}})Zhang, Chen, and Bui]{ehr:8}
Tianran Zhang, Muhao Chen, and Alex~AT Bui.
\newblock Diagnostic prediction with sequence-of-sets representation learning for clinical events.
\newblock In \emph{Artificial Intelligence in Medicine: 18th International Conference on Artificial Intelligence in Medicine, AIME 2020, Minneapolis, MN, USA, August 25--28, 2020, Proceedings 18}, pages 348--358. Springer, 2020{\natexlab{a}}.

\bibitem[Zhang et~al.(2019)Zhang, Tang, Dodge, Zhou, and Wang]{other:9}
Xi~Sheryl Zhang, Fengyi Tang, Hiroko~H Dodge, Jiayu Zhou, and Fei Wang.
\newblock Metapred: Meta-learning for clinical risk prediction with limited patient electronic health records.
\newblock In \emph{Proceedings of the 25th ACM SIGKDD international conference on knowledge discovery \& data mining}, 2019.

\bibitem[Zhang et~al.(2020{\natexlab{b}})Zhang, Qian, Cao, Li, Chen, Zheng, and Davidson]{ehr:12}
Xianli Zhang, Buyue Qian, Shilei Cao, Yang Li, Hang Chen, Yefeng Zheng, and Ian Davidson.
\newblock Inprem: An interpretable and trustworthy predictive model for healthcare.
\newblock In \emph{Proceedings of the 26th ACM SIGKDD International Conference on Knowledge Discovery \& Data Mining}, pages 450--460, 2020{\natexlab{b}}.

\bibitem[Zhao et~al.(2021)Zhao, Qiao, Xiao, Glass, and Sun]{other:11}
Yue Zhao, Zhi Qiao, Cao Xiao, Lucas Glass, and Jimeng Sun.
\newblock Pyhealth: A python library for health predictive models, 2021.

\end{thebibliography}

\appendix

\section{Deep Learning on Hypergraphs} \label{apd:first}
The success of graph neural networks \citep{other:1} has spurred hypergraph-based deep learning methods. Models updating node embeddings of hypergraphs through clique-expansion or its variants were proposed \citep{hyper:5, hyper:7, hyper:8, hyper:14}. HyperSAGE \citep{hyper:11} was developed for inductive hypergraph learning, capturing relations within and across hyperedges. HNHN \citep{hyper:13} employed a normalization strategy that could be adjusted according to datasets. \citet{hyper:12} innovated edge representation learning on graphs by introducing dual hypergraphs. HCNH \citep{hyper:3} utilized the hypergraph reconstruction loss for semi-supervised node classification. \citet{hyper:4} introduced AllSet, a generalized framework encapsulating most existing propagation rules on hypergraphs. For EHR-based tasks, HCL, ProCare, and CACHE emerged as three hypergraph neural network models \citep{hyper:1, tan2023enhancing, hyper:2}, with the former two lacking interpretability and the last one prone to false negative diagnoses.

\section{Comparing Hypergraph Neural Networks} \label{apd:second}
To optimize the performance of \method in diagnosis prediction, we experimented with $8$ state-of-the-art hypergraph neural network architectures. As \method represents each patient through a distinct hypergraph and all patient hypergraphs utilize a shared message-passing mechanism, only inductive hypergraph neural network models can be considered. The results in \tableref{tab:hypergraphs} indicate that UniGIN \citep{using:8}, AllSetTransformer \citep{hyper:4}, and UniGAT outperform others on the MIMIC-III dataset, whereas UniGIN, UniGCN, and HyperGCN \citep{hyper:14} are the top performers on MIMIC-IV. Given UniGIN's consistently high performance across both datasets and its fewer parameters relative to UniGAT and AllSetTransformer, we adopted UniGIN as the message-passing mechanism for \method.

\begin{table*}[ht]
    \floatconts
    {tab:hypergraphs}
    {\caption{The Comparison of Different Hypergraph Neural Network Architectures}}
    {
    \begin{threeparttable}
        \resizebox{\textwidth}{!}{
            \begin{tabular}{@{}c|cccccccccc@{}}
                \toprule
                Model & Recall@$10$ & Recall@$20$ & nDCG@$10$ & nDCG@$20$ & \# Params & Recall@$10$ & \multicolumn{1}{l}{Recall@$20$} & \multicolumn{1}{l}{nDCG@$10$} & \multicolumn{1}{l}{nDCG@$20$} & \multicolumn{1}{l}{\# Params} \\ 
                \midrule
                & \multicolumn{5}{c}{MIMIC-III} & \multicolumn{5}{c}{MIMIC-IV} \\ 
                \midrule
                \method (UniGIN)            & \underline{$0.2775$} & \underline{$0.3831$} & \underline{$0.4135$} & $\mathbf{0.4088}$    & $2.69$M              & $\mathbf{0.3344}$    & $\mathbf{0.4270}$    & $\mathbf{0.4236}$    & $\mathbf{0.4239}$    & $3.59$M \\
                \method (UniSAGE)           & $0.2744$             & $0.3799$             & $0.4112$             & $0.4050$             & $2.69$M              & $0.3311$             & $0.4231$                         & $0.4205$             & $0.4199$             & $3.59$M \\
                \method (UniGAT)            & $\mathbf{0.2780}$    & $0.3828$             & $\mathbf{0.4141}$    & \underline{$0.4080$} & $2.70$M              &  $0.3321$            & $0.4240$                         & $0.4212$             & $0.4230$             & $3.59$M \\
                \method (UniGCN)            & $0.2758$             & $0.3735$             & $0.4132$             & $0.4055$             & $2.69$M              & \underline{$0.3338$} & \underline{$0.4268$} & $0.4208$             & \underline{$0.4232$} & $3.59$M \\
                \method (UniGCNII)          & $0.2659$             & $0.3722$             & $0.4098$             & $0.4009$             & $2.63$M              & $0.3328$             & $0.4251$                         & $0.4206$             & $0.4228$             & $3.53$M \\
                \method (HyperGCN)          & $0.2648$             & $0.3671$             & $0.4024$             & $0.3965$             & $2.50$M              & $0.3318$             & $0.4234$                         & \underline{$0.4216$} & $0.4229$             & $3.49$M \\
                \method (AllDeepSets)       & $0.2693$             & $0.3812$             & $0.4051$             & $0.4036$             & $3.08$M              & $0.3305$             & $0.4216$                         & $0.4200$             & $0.4215$             & $3.73$M \\
                \method (AllSetTransformer) & $0.2762$             & $\mathbf{0.3835}$    & $0.4123$             & \underline{$0.4080$} & $3.06$M              & $0.3303$             & $0.4224$                         & $0.4205$             & $0.4226$             & $3.73$M  \\
                \bottomrule
            \end{tabular}
        }
    \end{threeparttable}
    }
\end{table*}

\section{The GRU Decoder} \label{apd:third}
\method utilizes a GRU decoder to reconstruct patient hypergraphs from the embeddings of the temporal phenotypes, denoted as $[\mathbf{U}_{1}\mathbin\Vert\mathbf{U}_{2}\mathbin\Vert ... \mathbin\Vert\mathbf{U}_{K}] \in \mathbb{R}^{Kd_{hid}}$. Specifically, the reconstructed patient hypergraph, $\hat{\mathbf{P}}$, is derived as follows:
\begin{align*}
\mathbf{H}_{1}, ..., \mathbf{H}_{T} &= \textrm{GRU}([\mathbf{U}_{1}\mathbin\Vert ... \mathbin\Vert\mathbf{U}_{K}], ..., [\mathbf{U}_{1}\mathbin\Vert ... \mathbin\Vert\mathbf{U}_{K}])\\ 
\hat{\mathbf{P}} &= \sigma(\mathbf{H}\mathbf{W}_{\textrm{recon}} + \vec{b}_{\textrm{recon}})^{\!\!\top},
\end{align*}
where $\sigma$ denotes the sigmoid function, $\mathbf{H}_{t}$ is the hidden state corresponding to the $t$-th visit, and $\mathbf{W}_{\textrm{recon}}$ and $\vec{b}_{\textrm{recon}}$ are learnable parameters. In essence, \method duplicates the concatenated phenotype embedding $T$ times, providing the same input to the decoder GRU at each time step and reconstructing the columns of $\hat{\mathbf{P}}$ using the derived hidden states.

\section{More on Experimental Setup} \label{apd:fourth}
\subsection{Dataset Statistics}
Details on the employed datasets are in \tableref{tab:dataset_info}.

\begin{table}[hbtp]
\floatconts
  {tab:dataset_info}
  {\caption{Dataset Statistics}}
  {\resizebox{\columnwidth}{!}{\begin{tabular}{@{}l|cc@{}}
      \toprule
      Dataset                       & MIMIC-III & MIMIC-IV \\ \midrule
      \# of patients                & 7493      & 45250    \\
      \# of visits                  & 19894     & 162372   \\
      Avg. \# of visits per patient & 2.66      & 3.59     \\
      Max \# of visits per patient  & 42        & 95       \\
      Avg. \# of codes per visit    & 13.06     & 10.27    \\
      Max \# of codes per visit     & 39        & 39       \\
      \# of unique ICD-9 codes      & 4880      & 8402     \\ \bottomrule
    \end{tabular}}}
\end{table}

\subsection{Baseline Descriptions}
We compare \method with $13$ representative state-of-the-art models on diagnosis prediction:
\begin{itemize}[leftmargin=*]
  \item Doctor AI \citep{baseline:1} utilizes a GRU for learning patient representation.
  \item RETAIN \citep{baseline:2} employs RNNs and an attention mechanism for interpretable predictions.
  \item Dipole \citep{baseline:3} leverages bidirectional RNNs with an attention mechanism for interpretability.
  \item T-LSTM \citep{baseline:11} uses a time-aware long short-term memory to learn patient representation.
  \item GRAM \citep{baseline:6} infuses knowledge from medical ontologies into disease embeddings via attention.
  \item CGL \citep{baseline:7} achieves accurate diagnosis predictions via collaborative graph learning.
  \item Timeline \citep{baseline:12} ensures interpretability and prediction accuracy through learning time decay factors.
  \item AdaCare \citep{baseline:8} models correlations in diagnoses to retain interpretability and prediction accuracy.
  \item StageNet \citep{baseline:4} extracts disease stage information for accurate diagnosis prediction.
  \item ConCare \citep{baseline:9} models feature interactions with attention for personalized diagnosis prediction.
  \item Chet \citep{baseline:10} exploits transition functions on dynamic graphs to model disease progressions.
  \item SETOR \citep{baseline:13} employs neural ordinary differential equations for learning patient representation.
  \item MIPO \citep{baseline:15} leverages Transformer and medical ontologies for accurate diagnosis prediction.
\end{itemize}
For a fair comparison, while CGL utilizes unstructured texts in the EHR dataset to enhance prediction, we have opted to use it without its textual feature modeling module. Baseline models are implemented using the original authors' code or through PyHealth \citep{other:11}.

\subsection{Hyperparameters}
The datasets are split into training, validation, and testing sets using an $0.8$:$0.1$:$0.1$ ratio. We employ the Adam optimizer and standardize the batch size to $128$ across all models. We carefully tune the learning rate and specific hyperparameters of the baseline models to optimize their performance. Through grid search, hyperparameters for \method are finalized: $K=5$ and $Z=2$ for both MIMIC-III and MIMIC-IV. Details for other hyperparameters are available in our repository. The models are trained on a server with NVIDIA A40 GPUs. For all experiments, we present the average results from $5$ runs with random model initializations.

\section{More on Case Study} \label{apd:fifth}
\textbf{\method enhanced its explanatory capability by including false negatives.} For Phenotype $1$, a diagnosis of pneumonia was retrospectively added to the initial visit, reflecting the patient's respiratory problems, such as bronchiectasis, in the same visit, and subsequent pneumonia diagnoses; also, this adjustment further emphasized the patient’s enduring respiratory and immune system problems.

A clinician observed that \method initially overlooked corticoadrenal insufficiency, and none of the phenotypes included this diagnosis in the third visit. Since rheumatoid arthritis is an autoimmune disease, which could be a major cause of corticoadrenal insufficiency, this clinician decided to intervene by incorporating corticoadrenal insufficiency in the third visit into Phenotype $5$ and proceed with the predictive process using the adjusted phenotypes. Interestingly, this intervention allowed \method to accurately predict corticoadrenal insufficiency in the next visit, demonstrating its superior flexibility by allowing explicit interventions from domain experts in the prediction process.

\end{document}